\documentclass[runningheads]{llncs}
\usepackage{graphicx}
\usepackage{amsmath,amssymb} % define this before the line numbering.
\usepackage{color}

\usepackage[width=122mm,left=12mm,paperwidth=146mm,height=193mm,top=12mm,paperheight=217mm]{geometry}
\graphicspath{{MyFigures/}}
\usepackage{caption}
\usepackage{subcaption}
\usepackage{array}
\usepackage{multirow}
\usepackage[colorlinks,linkcolor=red,urlcolor=red,citecolor=green]{hyperref}
\begin{document}
\title{Reverse Attention for Salient Object Detection} 
% Replace with your title

\titlerunning{Reverse Attention for Salient Object Detection}
% Replace with a meaningful short version of your title
%
\author{Shuhan Chen \and
Xiuli Tan \and
Ben Wang \and
Xuelong Hu}
%
%Please write out author names in full in the paper, i.e. full given and family names. 
%If any authors have names that can be parsed into FirstName LastName in multiple ways, please include the correct parsing, in a comment to the volume editors:
%\index{Lastnames, Firstnames}
%(Do not uncomment it, because you may introduce extra index items if you do that, we will use scripts for introducing index entries...)
\authorrunning{Shuhan Chen et al.}
% Replace with shorter version of the author list. If there are more authors than fits a line, please use A. Author et al.
%

\institute{School of Information Engineering,\\
	Yangzhou University, China\\
	\email{\{c.shuhan, t.xiuli0214\}@gmail.com, wangben9503@163.com, xlhu@yzu.edu.cn}
}
\maketitle              % typeset the header of the contribution
\begin{abstract}
Benefit from the quick development of deep learning techniques, salient object detection has achieved remarkable progresses recently. However, there still exists following two major challenges that hinder its application in embedded devices, low resolution output and heavy model weight. To this end, this paper presents an accurate yet compact deep network for efficient salient object detection. More specifically, given a coarse saliency prediction in the deepest layer, we first employ residual learning to learn side-output residual features for saliency refinement, which can be achieved with very limited convolutional parameters while keep accuracy. Secondly, we further propose reverse attention to guide such side-output residual learning in a top-down manner. By erasing the current predicted salient regions from side-output features, the network can eventually explore the missing object parts and details which results in high resolution and accuracy. Experiments on six benchmark datasets demonstrate that the proposed approach compares favorably against state-of-the-art methods, and with advantages in terms of simplicity, efficiency (\textbf{45 FPS}) and model size (\textbf{81 MB}).

\end{abstract}
\section{Introduction}
Salient object detection, also known as saliency detection, aims to localize and segment the most conspicuous and eye-attracting objects or regions in an image. It is usually served as a pre-processing step to facilitate various subsequent high-level vision tasks, such as image segmentation~\cite{Wei1}, image captioning~\cite{Xu2}, and so on. Recently, with the quick development of deep convolutional neural networks (CNNs), salient object detection has achieved significant improvements over conventional hand-crafted feature based approaches. The emergence of fully convolutional neural networks (FCNs)~\cite{Long3} further pushed it to a new state-of-the-art due to its efficiency and end-to-end training. Such architecture also benefits other applications, \emph{e.g.}, semantic segmentation~\cite{Dai4}, edge detection~\cite{Xie5}.

\begin{figure}  
  \centering  
  \includegraphics[width=65mm]{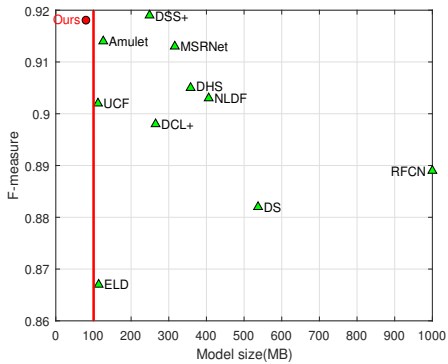}
  \caption{Maximum F-Measure of recent deep CNN-based saliency detection models on ECSSD, including DS~\cite{Li41}, ELD~\cite{Lee42}, DCL$^{+}$~\cite{Li6}, DHS~\cite{Li6}, RFCN~\cite{Wang21}, NLDF~\cite{Luo22}, DSS$^{+}$~\cite{Hou10}, MSRNet~\cite{Li9}, Amulet~\cite{Zhang14}, UCF~\cite{Zhang23}, and ours (red circle). As can be seen that the proposed model is the only one less than 100 MB while achieves comparable performance with state-of-the-art methods.}  
  \label{fig_size}  
\end{figure} 

Albeit profound progresses have been made, there still exists two major challenges that hinder its applications in real-world, \emph{e.g.}, embedded devices. One is the low resolution of the saliency maps produced by FCNs based saliency models. Due to the repeated stride and pooling operations in CNN architectures, it is inevitable to lose resolution and difficult to refine, making it infeasible to locate salient objects accurately, especially for the object boundaries and small objects. The other is the heavy weight and large redundancy of the existing deep saliency models. As can be seen in Fig.~\ref{fig_size}, all the listed deep models are larger than 100 MB, which is too heavy for a pre-processing step to apply in subsequent high-level tasks, and also not memory efficient for embedded devices.

Diverse solutions have been explored to improve the resolution of the FCNs based prediction. Early works~\cite{Li6,Chen8,Tang7} usually combined it with an extra region or superpixel based stream to fuse their respective advantages at the expense of high time cost. Then, some simple yet effective structures are constructed to combine the complementary cues of shallow and deep CNN features, which capture low-level spatial details and high-level semantic information respectively, such as skip connections~\cite{Li9}, short connections~\cite{Hou10}, dense connections~\cite{Xiao11}, adaptive aggregation~\cite{Zhang14}. Such multi-level feature fusion schemes also play an important role in semantic segmentation~\cite{Ronneberger46,Pinheiro15}, edge detection~\cite{Liu16}, skeleton detection~\cite{Ke18,Shen17}. Nevertheless, the existing archaic fusions are still incompetent for saliency detection under complex real-world scenarios, especially when dealing with multiple salient objects with diverse scales. In addition, some time consuming post-processing skills are also applied for refinement, \emph{e.g.}, superpixel-based filter~\cite{Hu12}, fully connected conditional random field (CRF)~\cite{Li6,Hou10,Krahenbuhl13}. However, to the best of our knowledge, there are no saliency detection networks explored considering both lightweight model and high accuracy.

\begin{figure}
\captionsetup[subfigure]{labelformat=empty}	
	\centering
       \begin{subfigure}[t]{1.75cm}
		\centering
		\includegraphics[width=1.75cm]{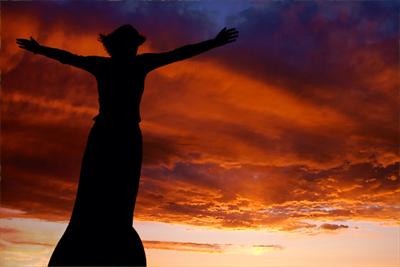}
	\end{subfigure}
	\begin{subfigure}[t]{1.75cm}
		\centering
		\includegraphics[width=1.75cm]{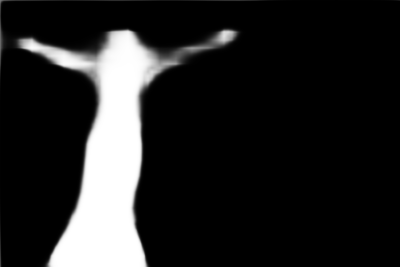}
	\end{subfigure}
	\begin{subfigure}[t]{1.75cm}
		\centering
		\includegraphics[width=1.75cm]{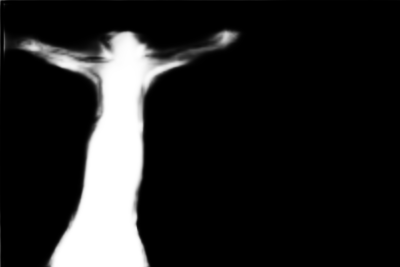}
	\end{subfigure}
	\begin{subfigure}[t]{1.75cm}
		\centering
		\includegraphics[width=1.75cm]{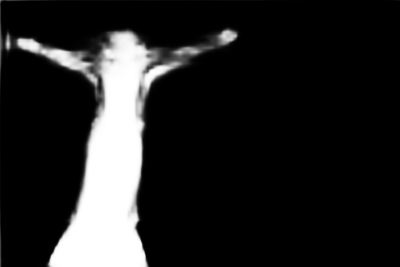}
	\end{subfigure}
	\begin{subfigure}[t]{1.75cm}
		\centering
		\includegraphics[width=1.75cm]{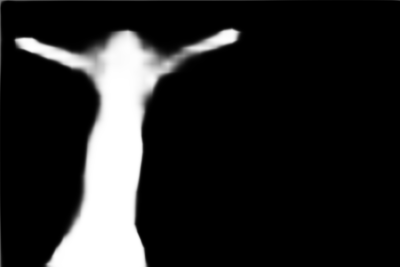}
	\end{subfigure}
	\begin{subfigure}[t]{1.75cm}
		\centering
		\includegraphics[width=1.75cm]{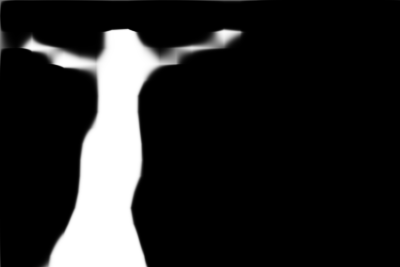}
	\end{subfigure}
        
        \begin{subfigure}[t]{1.75cm}
		\centering
		\includegraphics[width=1.75cm]{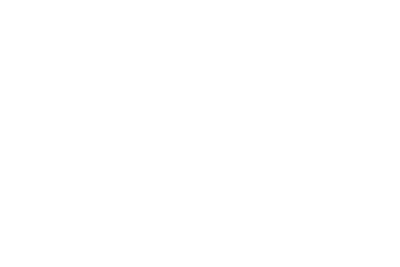}
	\end{subfigure}
	\begin{subfigure}[t]{1.75cm}
		\centering
		\includegraphics[width=1.75cm]{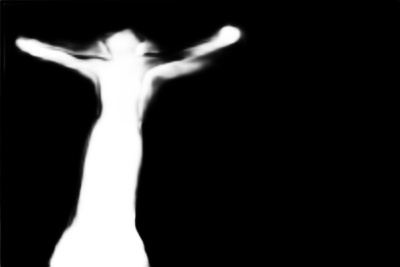}
	\end{subfigure}
	\begin{subfigure}[t]{1.75cm}
		\centering
		\includegraphics[width=1.75cm]{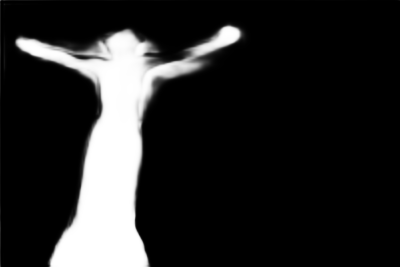}
	\end{subfigure}
	\begin{subfigure}[t]{1.75cm}
		\centering
		\includegraphics[width=1.75cm]{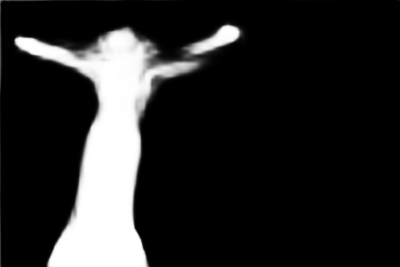}
	\end{subfigure}
	\begin{subfigure}[t]{1.75cm}
		\centering
		\includegraphics[width=1.75cm]{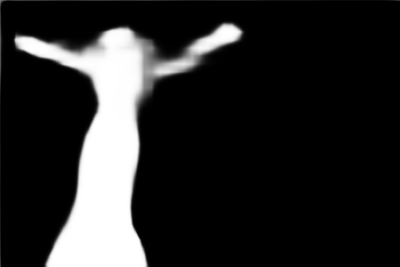}
	\end{subfigure}
	\begin{subfigure}[t]{1.75cm}
		\centering
		\includegraphics[width=1.75cm]{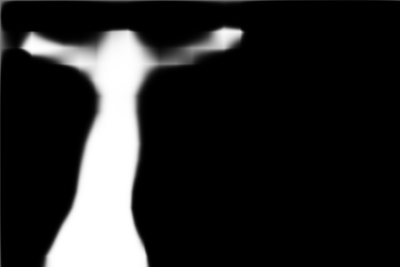}
	\end{subfigure}
        
        \begin{subfigure}[t]{1.75cm}
		\centering
		\includegraphics[width=1.75cm]{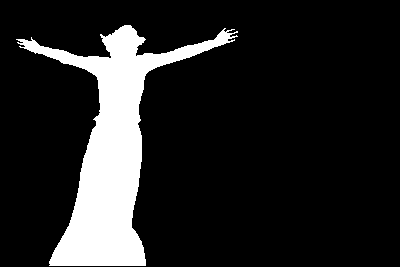}
		\caption{\scriptsize Img$\&$GT}	
	\end{subfigure}
	\begin{subfigure}[t]{1.75cm}
		\centering
		\includegraphics[width=1.75cm]{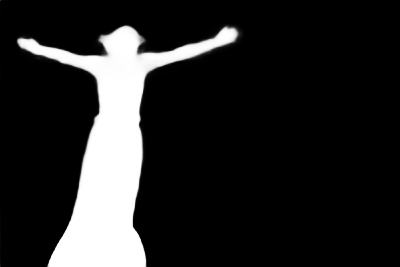}
		\caption{\scriptsize s-out 1}	
	\end{subfigure}
	\begin{subfigure}[t]{1.75cm}
		\centering
		\includegraphics[width=1.75cm]{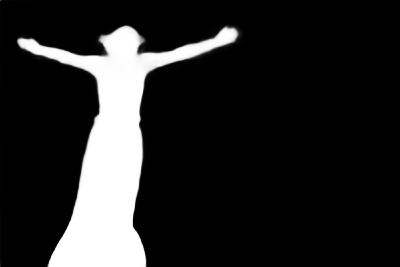}
		\caption{\scriptsize s-out 2}
	\end{subfigure}
	\begin{subfigure}[t]{1.75cm}
		\centering
		\includegraphics[width=1.75cm]{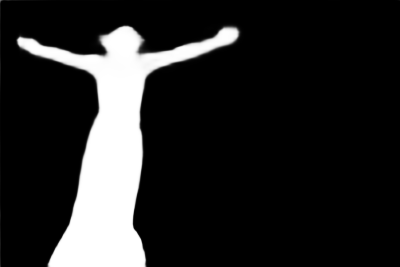}
		\caption{\scriptsize s-out 3}
	\end{subfigure}
	\begin{subfigure}[t]{1.75cm}
		\centering
		\includegraphics[width=1.75cm]{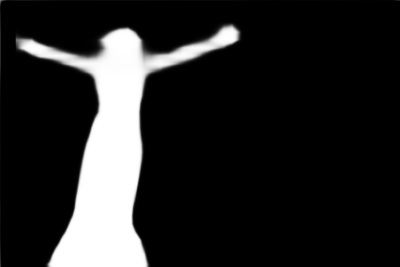}
		\caption{\scriptsize s-out 4}
	\end{subfigure}
	\begin{subfigure}[t]{1.75cm}
		\centering
		\includegraphics[width=1.75cm]{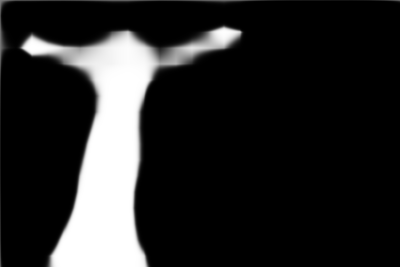}
		\caption{\scriptsize s-out 5}
	\end{subfigure} 
        \caption{Visual comparison of saliency maps produced by DSS~\cite{Hou10} (top row), our method without (middle row) and with reverse attention (bottom row) in different side-outputs, respectively. As can be seen clearly that the resolutions of the saliency maps are improved gradually from deep to shallow side-outputs, and our reverse attention based side-output residual learning performs much better than short connections~\cite{Hou10}.}\label{fig_sout}
\end{figure}

To this end, we present an accurate yet compact deep salient object detection network which achieved comparable performance with state-of-the-art methods, thus enables for real-time applications. In generally, more convolutional channels with large kernel size leads to better performance in salient object detection due to the large receptive field and model capacity to capture more semantic information, \emph{e.g.}, there are 512 channels with kernel size 7$\times$7 in the last side-output of DSS~\cite{Hou10}. In a different way, we introduce residual learning~\cite{He19} into the architecture of HED~\cite{Xie5}, and regard salient object detection as a super-resolution reconstruction problem~\cite{Lai44}. Given the low resolution prediction of FCNs, side-output residual features are learned to refine it step by step. Note that it can be achieved only using convolution with 64 channels and kernel size 3$\times$3 in each side-output, whose parameters are significant fewer than DSS. 

Similar residual learning was also utilized in skeleton detection~\cite{Ke18} and image super-resolution~\cite{Kim25}. However, the performance is not satisfactory enough if we directly apply it for salient object detection due to its challenging. Since most of the existing deep saliency models are fine-tuned from image classification network, the fine-tuned network will unconsciously focus on the regions with high response values during residual learning as can be seen in Fig.~\ref{fig_feature}, thus struggling to capture the residual details, \emph{e.g.}, object boundaries and other undetected object parts. To solve it, we propose reverse attention to guide side-output residual learning in a top-down manner. Specifically, prediction of deep layer is upsampled then reversed to weight its neighbor shallow side-output feature, which quickly guides the network to focus on the undetected regions for residual capture, thus leads to better performance as seen in Fig.~\ref{fig_sout}.

In summary, the contributions of this paper can be concluded as: (1) We introduce residual learning into the architecture of HED for salient object detection. With the help of the learned side-output residual features, the resolution of the saliency map can be improved gradually with much fewer parameters compared to the existing deep saliency networks. (2) We further propose reverse attention to guide side-output residual learning. By erasing the current prediction, the network can discover the missing object parts and residual details effectively and quickly, which leads to significant performance improvement. (3) Benefit from the above two components, our approach consistently achieves comparable performance with state-of-the-art methods, and with advantages in terms of simplicity, efficiency (45 FPS) and model size (81 MB).

\section{Related Work}

There are plenty of saliency detection methods proposed in the past two deceads. Here, we only focus on the recent state-of-the-art methods. Almost all of them are FCNs based and try to solve the common problem: how to produce saliency map with high resolution by using FCNs? Kuen \emph{et al.}~\cite{Kuen20} applied recurrent unit into FCNs to iteratively refine each salient region. Hu \emph{et al.}~\cite{Hu12} entended a superpixel-based guided filter to be a layer in the network for boundary refinement. Hou \emph{et al.}~\cite{Hou10} designed short connections for multi-scale feature fusion, while in Amulet~\cite{Zhang14}, multi-level convolutional features were aggregated adaptively. Luo \emph{et al.}~\cite{Luo22} proposed a multi-resolution grid structure to capture both local and global cues. In addition, a new loss function was introduced to penalize errors on the boundaries. Zhang \emph{et al.}~\cite{Zhang23} further proposed a novel upsampling method to reduce the artifacts produced in deconvolution. Recently, dilated convolution~\cite{Hu12} and dense connections~\cite{Xiao11} are further incorporated to obtain high resolution saliency map. There are also some progressive works to address the above issue in semantic segmentation. In~\cite{Pinheiro15}, skip connections was proposed to refine object instances, while in~\cite{Ghiasi24}, it was used to build a Laplacian pyramid reconstruction network for object boundary refinement.

Instead of fusing multi-level convolutional features as the above works, we try to learn residual feature for low resolution refinement. The idea of residual learning was first proposed by He \emph{et al.}~\cite{He19}for image classification. After that, it was widely applied in various applications. Ke \emph{et al.}~\cite{Ke18} leraned side-output residual feature for accurate object symmetry detection. Kim \emph{et al.}~\cite{Kim25} built a very deep convolutional network based on residual learning for accurate image super-resolution.

Although it is natural to apply it for salient object detection, the performance is not satisfactory enough. To solve it, we introduce attention mechanism which is inspired from human perception process. By using top information to efficiently guide bottom-up feedforward process, it has achieved great success in many tasks. Attention model was designed to weight multi-scale features in~\cite{Li9,Chen26}. Residual attention module was stacted to generate deep attention-aware features for image classification in~\cite{Wang27}. In ILSVRC 2017 Image Classification Challenge, Hu \emph{et al.}~\cite{Hu28} won the 1st place by constructing “Squeeze-and-Excitation” block for channel attention. Huang \emph{et al.}~\cite{Huang45} designed an attention mask to highlight the prediction of the reverse object class, which then be subtracted from the original prediction to correct the mistakes in the confusion area for semantic segmentation. Inspired but differed from it, we employ reverse attention in a top-down manner to guide side-output residual learning. Benefit from it, we can learn more accurate residual details which leads to significant improvement.

\begin{figure}  
  \centering  
  \includegraphics[width=110mm]{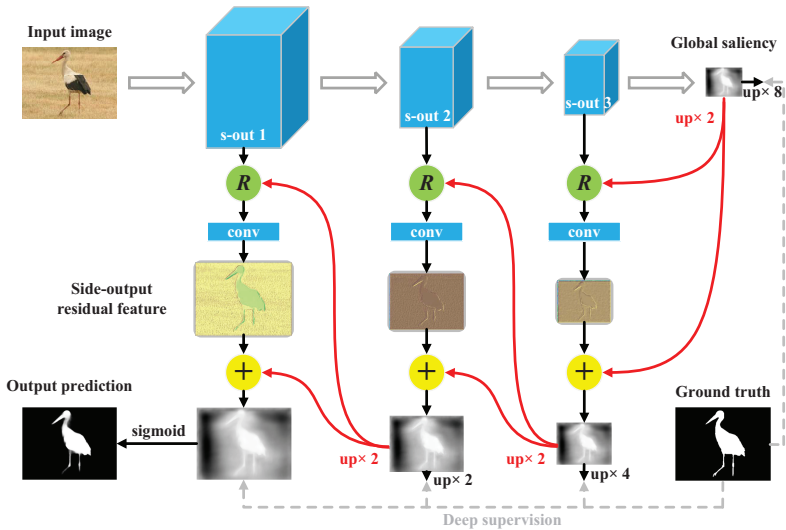}
  \caption{The overall architecture of the proposed network. Here, only three side-outputs are listed for illustration.``$R$'' denotes the proposed reverse attention block that is illustrated in Fig.~\ref{fig_ra}. As can be seen, the residual error decreases along the stacking orientation with the supervision both on the input and output of the residual unit (yellow circle).}  
  \label{fig_architecture}  
\end{figure} 

\section{Proposed Method}

In this section, we first describe the overall architecture of the proposed deep salient object detection network, and then present the details of the main components one by one, which are corresponding to side-output residual learning and top-down reverse attention respectively.

\subsection{Architecture}

The proposed network is built upon the HED~\cite{Xie5} architecture and choses VGG-16~\cite{Simonyan29} as backbone. We use the layers up to ``pool5'' and select \{conv1\_2, conv2\_2, conv3\_3, conv4\_3, conv5\_3\} as side-outputs, which have strides of \{1, 2, 4, 8, 16\} pixels with respect to the input image respectively. We first reduce the dimension of ``pool5'' into 256 by convolution with kernel size 1$\times1$, and then add three convolutional layers with 5$\times5$ kernels to capture global saliency. Since the resolution of the global saliency map is only 1/32 of the input image, we further learn residual feature in each side-output to improve its resolution gradually. In specifically, $D$ convolutional layers with 3$\times3$ kernels and 64 channels are stacked for residual learning. The reverse attention block is embedded before side-output residual learning. The prediction of the shallowest side-output is fed into a sigmoid layer for final output. The overall architecture is shown in Fig.~\ref{fig_architecture} and complete configurations are outlined in Table~\ref{table_setting}.

\begin{table}[]
\centering
\caption{The configurations of the proposed network. $(n, k\times k)\times D$ denotes stacking $D$ convolutional layers with channel number ($n$) and kernel size ($k$), and ReLU layer is added for nonlinear transformation.}
\label{table_setting}
\begin{tabular}{p{4.2cm}<{\centering} p{4.2cm}<{\centering}}
\hline
Side output 1$\sim$5 & Global saliency \\
\hline
(64, 1$\times$1) & (256, 1$\times$1)  \\
$\left\{ {(64, 3\times3),\rm ReLU} \right\}\times D$ & $\left\{ {(256, 5\times5),\rm ReLU} \right\}\times3$  \\
(1, 3$\times$3) & (1, 1$\times$1)  \\
\hline
\end{tabular}
\end{table}

\subsection{Side-output Residual Learning}

As we know, deep layers of network capture high-level semantic information but messy details, while it is opposite for shallow ones. Based on this observation, multi-level features fusion is a common choice to capture their complementary cues, however, it will degrade the confident prediction of deep layers when combining with shallow ones. In this paper, we implement it in a different yet more efficient way by employing residual learning to remedy the errors between the predicted saliency maps and the ground truth. Specifically, the residual feature is learned by applying deep supervision both on the input and output of the designed residual unit, which is illustrated in Fig.~\ref{fig_architecture}. Formally, given the upsampled input saliency map $S_{i+1}^{up}$ by a factor 2 in side-output stage $i+1$, and the residual feature $R_{i}$ learned in side-output stage $i$, then the deep supervision can be formulated as:

\begin{equation}
\left\{
\begin{aligned}
& \{S_{i+1}\}^{up\times2^{i+1}} \approx G \\
\{S_{i+1}^{up}&+R_{i}\}^{up\times2^{i}}=\{S_{i}\}^{up\times2^{i}}\approx G \\
\end{aligned}
\quad,\right.
\end{equation}
where $S_{i}$ is the output of the residual unit and $G$ is ground truth, $up\times2^{i}$ denotes the upsample operation by a factor $2^{i}$, which is implemented by the same bilinear interpolation with HED~\cite{Xie5}.

Such a learning objective inherits the following good property. The residual units establish shortcut connections between the predictions from different scales and the ground truth, which makes it easier to remedy their errors with higher scale adaptability. Generally, the error between the input and output of the residual unit is fairly small based on the same supervision, thus can be learned more easily with fewer parameters and iterations. To the extreme, the error is approximately equal to zero if the prediction is close enough to the ground truth. As a result, the constructed network can be very efficient and lightweight.

\subsection{Top-down Reverse Attention}

\begin{figure}  
  \centering  
  \includegraphics[width=110mm]{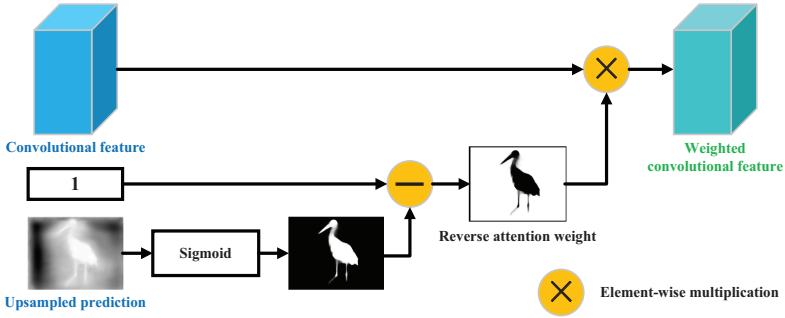}
  \caption{Illustration of the proposed reverse attention block, whose input and output are highlighted in blue and green respectively.}  
  \label{fig_ra}  
\end{figure} 

Although it is natural and straightforward to learn residual details for saliency refinement, it is not easy for the network to capture them accurately without extra supervision, which will result in unsatisfactory detection. Since most of the existing saliency detection networks are fine-tuned from image classification networks which are only responsive to small and sparse discriminative object parts, it obviously deviates from the requirement of the saliency detection task that needs to explore dense and integral regions for pixel-wise prediction. To mitigate this gap, we propose a reverse attention based side-output residual learning approach for expanding object regions progressively. Starting with a coarse saliency map generated in the deepest layer with high semantic confidence but low resolution, our proposed approach guides the whole network to sequentially discover complement object regions and details by erasing the current predicted salient regions from side-output features, where the current prediction is upsampled from its deeper layer. Such a top-down erasing manner can eventually refine the coarse and low resolution prediction into a complete and high resolution saliency map with these explored regions and details, see Fig.~\ref{fig_ra} for illustration.

Given the side-output feature $T$ and reverse attention weight $A$, then the output attentive feature can be produced by their element-wise multiplication, which can be formulated as:

\begin{equation}
F_{z,c}=A_{z} \cdot T_{z,c},
\end{equation}
where $z$ and $c$ denote the spatial position of the feature map and the index of the feature channel, respectively. And the reverse attention weight in side-output stage $i$ is simply generated by subtracting the upsampled prediction of side-output $i+1$ from one, which is computed as below:

\begin{equation}
A_{i}=1-{\rm Sigmoid}(S_{i+1}^{up}).
\end{equation}

\begin{figure}
\captionsetup[subfigure]{labelformat=empty}	
	\centering
       \begin{subfigure}[t]{0.225\linewidth}
		\centering
		\includegraphics[height=2.05cm]{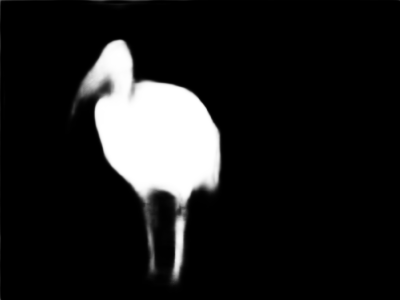}	
	\end{subfigure}
      \begin{subfigure}[t]{2.05cm}
		\centering
		\includegraphics[height=2.05cm]{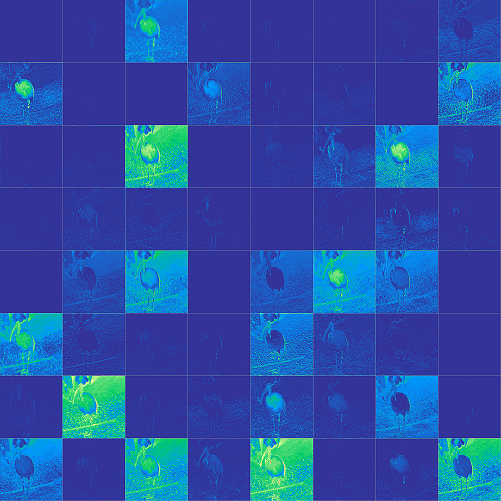}	
	\end{subfigure}
      \begin{subfigure}[t]{2.05cm}
		\centering
		\includegraphics[height=2.05cm]{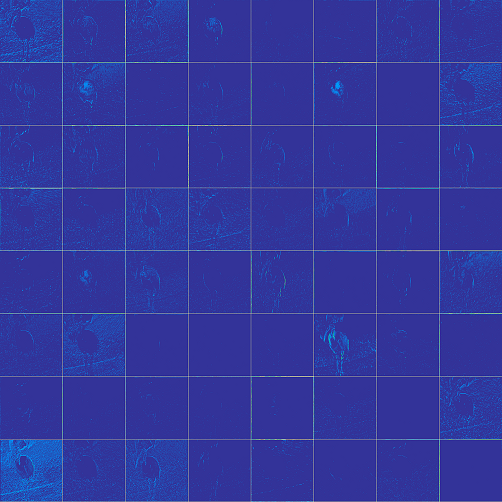}	
	\end{subfigure}
      \begin{subfigure}[t]{2.05cm}
		\centering
		\includegraphics[height=2.05cm]{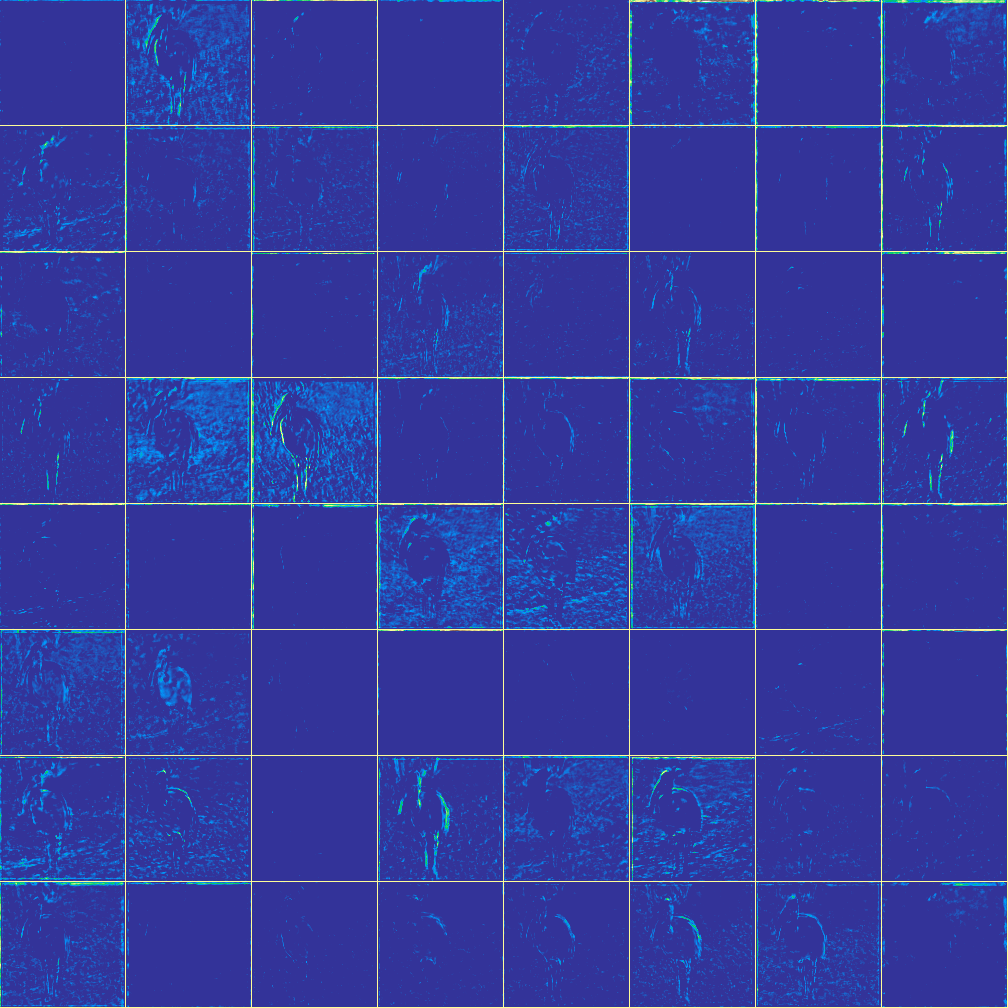}	
	\end{subfigure}
      \begin{subfigure}[t]{2.05cm}
		\centering
		\includegraphics[height=2.05cm]{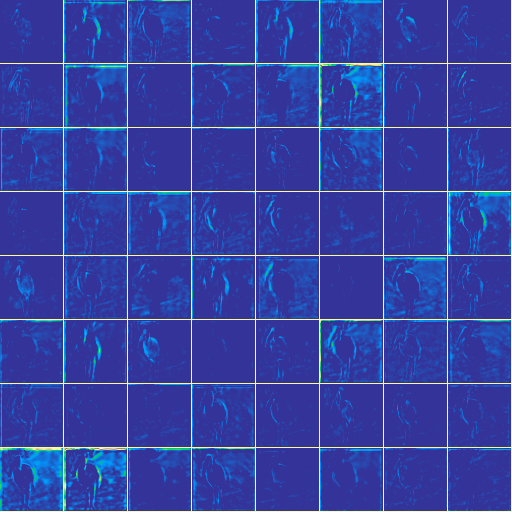}	
	\end{subfigure}

       \begin{subfigure}[t]{0.225\linewidth}
		\centering
		\includegraphics[height=2.05cm]{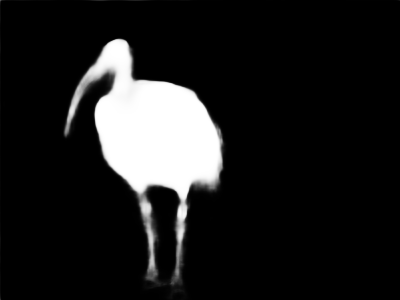}	
	\end{subfigure}
      \begin{subfigure}[t]{2.05cm}
		\centering
		\includegraphics[height=2.05cm]{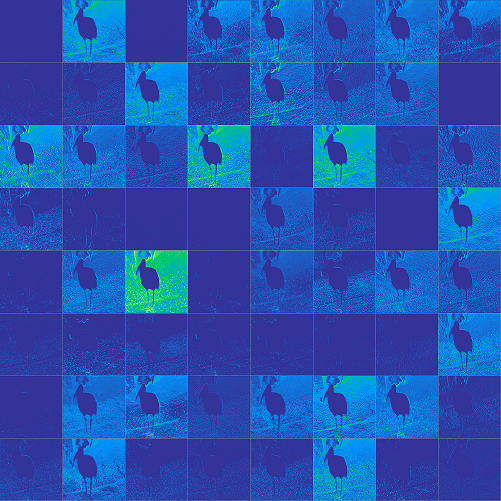}	
	\end{subfigure}
      \begin{subfigure}[t]{2.05cm}
		\centering
		\includegraphics[height=2.05cm]{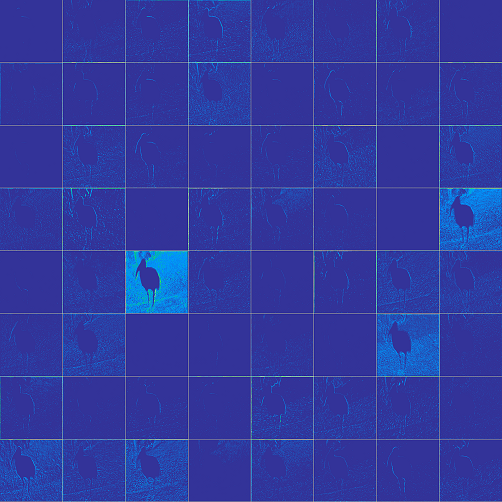}	
	\end{subfigure}
      \begin{subfigure}[t]{2.05cm}
		\centering
		\includegraphics[height=2.05cm]{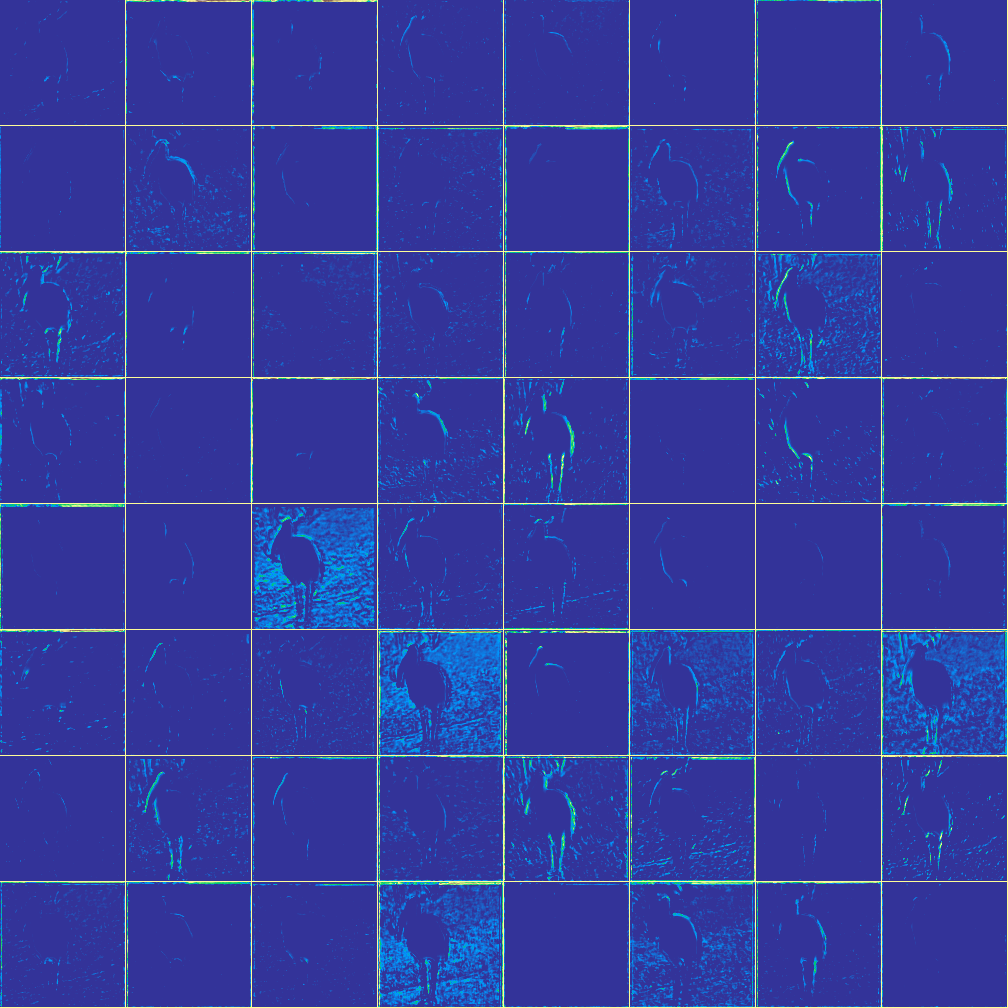}	
	\end{subfigure}
      \begin{subfigure}[t]{2.05cm}
		\centering
		\includegraphics[height=2.05cm]{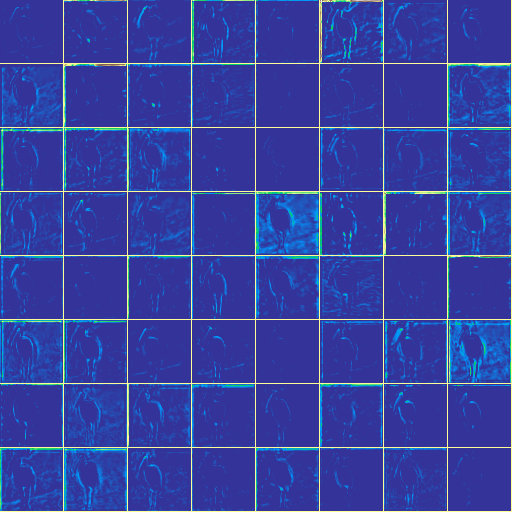}	
	\end{subfigure}
	\caption{Visualization of residual features in different side-outputs of the proposed network without (the first row) and with reverse attention (the second row). From left to right are saliency map, the last convolutional feature from side output 1 to 4, respectively. After applying our reverse attention, the proposed network well captured spatial details near object boundaries which is beneficial for saliency refinement, especially in shallow layers. Best viewed in color.}\label{fig_feature}
\end{figure}

Fig.~\ref{fig_feature} shows some visual examples of the learned residual feature to illustrate the effectiveness of the proposed reverse attention. As can be seen, the proposed network well captured the residual details near object boundaries with the help of reverse attention. While without reverse attention, it learned some redundant features inside object which is helpless for saliency refinement.

\subsection{Supervision}

As shown in Fig.~\ref{fig_architecture}, deep supervision is applied to each side-output stage as did in~\cite{Xie5,Hou10}. Each side-output produces a loss term $\mathcal{L}_{side}$ which is defined as below:

\begin{equation}
\mathcal{L}_{\rm side}(I,G,{\rm W},{\rm w})=\sum_{m=1}^{M}\ell_{\rm side}^{(m)}(I,G,{\rm W},{\rm w}^{(m)}),
\end{equation}
where $M$ regards to the total side-output numbers including global saliency, W denotes the collection of all standard network layer parameters, $I$ and $G$ refer to the input image and the corresponding ground truth respectively. Each side-output layer is regarded as a pixel-wise classifier with the corresponding weights w which is represented by

\begin{equation}
{\rm w}=({\rm w}^{(1)},{\rm w}^{(2)},...,{\rm w}^{(M)}).
\end{equation}
Here, $\ell_{\rm side}^{(m)}$ represents the image-level class-balanced cross-entropy loss function~\cite{Xie5} of the $m$th side output, which is computed by the following formulation:

\begin{equation}
\begin{split}
\ell_{\rm side}^{(m)}(I,G,{\rm W},{\rm w}^{(m)})=&-\sum_{z=1}^{|I|}G(z)logPr(G(z)=1|I(z);{\rm W},{\rm w}^{(m)})\\
&+(1-G(z))logPr(G(z)=0|I(z);{\rm W},{\rm w}^{(m)}),
\end{split}
\end{equation}
where $Pr(G(z)=1|I(z);W,w^{(m)})$ represents the probability of the activation value at location $z$ in the $m$th side output, $z$ is the spatial coordinate. Different with HED~\cite{Xie5} and DSS~\cite{Hou10}, there is no fusion layer included in our approach. The output of the first side-output is used as our final prediction after a sigmoid layer in the testing stage.

\subsection{Difference to Other Networks}

Though shares the same name, the proposed network significantly differs from reverse attention network~\cite{Huang45}, which applied reverse attention to weight the prediction that is not associated with a target class, in this way to amplify the reverse-class response in the confused region, thus can help the original branch make correct prediction. While in our approach, the usage of reverse attention is totally different. It is used to erase the confident prediction from deep layer, which can guide the network to explore the missing object regions and details effectively.

There are also some significant differences with other residual learning based architectures, \emph{e.g.}, side-output residual network (SRN)~\cite{Ke18}, and Laplacian reconstruction network (LRN)~\cite{Ghiasi24}. In SRN, the residual feature is learned from each side-output of VGG-16 directly, while in this paper, it is learned after reverse attention that is applied to guide residual learning. The main difference with LRN lies in the usage of the weight mask, which is used to weight the learned side-output features for boundary refinement in LRN, in contrast, we apply it before side-output feature learning for guidance. In addition, the weight mask in LRN is generated from the edge of deep prediction which will miss some object regions due to its low resolution, while in this paper, we apply it to focus on all the undetected regions for saliency refinement, which not only refines object boundaries well but also highlights object regions more completely. 

\section{Experiments}

\subsection{Experimental Setup}

The proposed network is built on the top of the implementations of HED~\cite{Xie5} and DSS~\cite{Hou10}, and trained though the publicly available Caffe~\cite{Jia30} library. The whole network is trained end-to-end using full-resolution images and optimized by stochastic gradient descent method. The hyper-parameters are set as below: batch size (1), iter\_size (10), the momentum (0.9), the weight decay (5e-4), learning rate is initialized as 1e-8 and decreased by 10\% when the training loss reaches a flat, the training iteration number (10K). All these parameters were fixed during the following experiments. The source code will be released\footnote{\url{http://shuhanchen.net}}.

We comprehensively evaluated our method on six representative datasets, including MSRA-B~\cite{Liu31}, HKU-IS~\cite{Li32}, ECSSD~\cite{Shi33}, PASCAL-S~\cite{Li34}, SOD~\cite{Martin35}, and DUT-OMRON~\cite{Yang36}, which contain 5000, 4447, 1000, 850, 300, 5168 well annotated images, respectively. Among them, PASCAL-S and DUT-OMRON are more challenging than the others. To guarantee a fair comparison with the existing approaches, we utilize the same training sets as in~\cite{Li6,Luo22,Hou10,Jiang37} and test all of the datasets with the same model. Data augmentation is also implemented the same with~\cite{Luo22,Hou10} to reduce the over-fitting risk, which increased by 2 times through horizontal flipping.

Three standard and widely agreed metrics are used to evaluate the performance, including Precision-Recall (PR) curve, F-measure, and the Mean Absolute Error (MAE). Pairs of precision and recall values are calculated by comparing the binary saliency maps with the ground truth to plot the PR curve, where the thresholds are in the range of [0, 255]. The F-measure is adopted to measure the overall performance, which is defined as the weighted harmonic mean of precision and recall:

\begin{equation}
F_{\beta}=(1+\beta^{2})\frac{{\rm Precision}\times{\rm Recall}}{{\beta^{2}}{\rm Precision}+{\rm Recall}},
\end{equation}
where $\beta^{2}$ is set to 0.3 to emphasize the precision over recall as suggested in~\cite{Borji38}. Only the maximum F-Measure is reported here to show the best performance a detector can achieve. Given the normalized saliency map $S$ and ground truth $G$, the MAE score is calculated by their average per-pixel difference:

\begin{equation}
{\rm MAE}=\frac{1}{H\times W} \sum_{x=1}^{H}\sum_{y=1}^{W}\left| S(x,y)-G(x,y) \right|,
\end{equation} 
where $W$ and $H$ are the width and height of the saliency map, respectively.

\subsection{Ablation Studies}

Before comparing with the state-of-the-art methods, we first evaluate the influence of different design options (the depth $D$), the effectiveness of the proposed side-output residual learning and reverse attention in this section. 

\textbf{Depth \emph{D}}. We make an experiment to see how the depth $D$ affects the performance by varying it from 1 to 3. The results on PASCAL-S and DUT-OMRON are shown in Table~\ref{table_d}. As can be seen that the best performance is obtained when $D$=2. Therefore, we set it as 2 in the following experiments.

\begin{table}[]
\centering
\caption{Performance comparison with different numbers of $D$.}
\label{table_d}
\begin{tabular}{p{1.5cm}<{\centering} p{1.5cm}<{\centering} p{1.5cm}<{\centering} p{1.5cm}<{\centering} p{1.5cm}<{\centering}}
\\\hline
                                & \multicolumn{2}{c}{PASCAL-S} & \multicolumn{2}{c}{DUT-OMRON} \\
                                 \cline{2-5}  & $F_{\beta}$ & MAE & $F_{\beta}$ & MAE \\\hline
$D$=1 & 0.830 & \textbf{0.100} & 0.776 & 0.067 \\
$D$=2 & \textbf{0.834} & 0.104 & \textbf{0.786} & \textbf{0.062} \\
$D$=3 & 0.824 & 0.106 & 0.778 & 0.064 \\\hline                                 
\end{tabular}
\end{table}

\textbf{Side-output residual learning}. To investigate the effectiveness of the side-output residual learning, we separately evaluate the performance of each side-output prediction and show in Table~\ref{table_rsl}. We can find that the performance is gradually improved by combing more side-output residual features.

\begin{table}[]
\centering
\caption{Performance comparison with different side-output predictions.}
\label{table_rsl}
\begin{tabular}{p{2.8cm}<{\centering} p{1.5cm}<{\centering} p{1.5cm}<{\centering} p{1.5cm}<{\centering} p{1.5cm}<{\centering}}
\\\hline
                                & \multicolumn{2}{c}{PASCAL-S} & \multicolumn{2}{c}{DUT-OMRON} \\
                                \cline{2-5}  & $F_{\beta}$ & MAE & $F_{\beta}$ & MAE \\\hline
Side-output 5 & 0.817 & 0.111 & 0.755 & 0.071 \\
Side-output 4 & 0.827 & 0.106 & 0.776 & 0.065 \\
Side-output 3 & 0.831 & \textbf{0.104} & 0.785 & \textbf{0.062} \\
Side-output 2 & 0.832 & \textbf{0.104} & \textbf{0.786} & \textbf{0.062} \\
Side-output 1 & \textbf{0.834} & \textbf{0.104} & \textbf{0.786} & \textbf{0.062} \\\hline                                 
\end{tabular}
\end{table}

\textbf{Reverse attention}. As illustrated in Fig.~\ref{fig_feature}, the network well located at the object boundaries with the help of reverse attention. Here, we perform a detailed comparison using F-measure and MAE scores which are reported in Table~\ref{table_results}. From the results, we can get the following observations: (1) Without reverse attention, our performance is similar to the state-of-the-art method DSS (without CRF-based post-processing), which indicates its large redundancy. (2) After applying reverse attention, the performance is improved by a large margin, specifically, we obtained an average of 1.4\% gain in terms of F-measure and 0.5\% decrease for MAE score, which clearly demonstrates its effectiveness.
                                
\subsection{Performance Comparison with State-of-the-art}

We compare the proposed method with 10 state-of-the-art ones, including 9 recent CNN-based approaches, DCL$^{+}$~\cite{Li6}, DHS~\cite{Liu39}, SSD~\cite{Kim40}, RFCN~\cite{Wang21}, DLS~\cite{Hu12}, NLDF~\cite{Luo22}, DSS and DSS$^{+}$~\cite{Hou10}, Amulet~\cite{Zhang14}, UCF~\cite{Zhang23}, and one conventional top approach, DRFI~\cite{Jiang37}, where symbol ``+'' indicates that the network includes CRF-based post-processing. Note that all the saliency maps of the above methods are produced by running source codes or pre-computed by the authors, and ResNet based methods are not included for fair comparison.

\begin{figure}
	\centering
	\begin{subfigure}[t]{4cm}
		\centering
		\includegraphics[width=4cm]{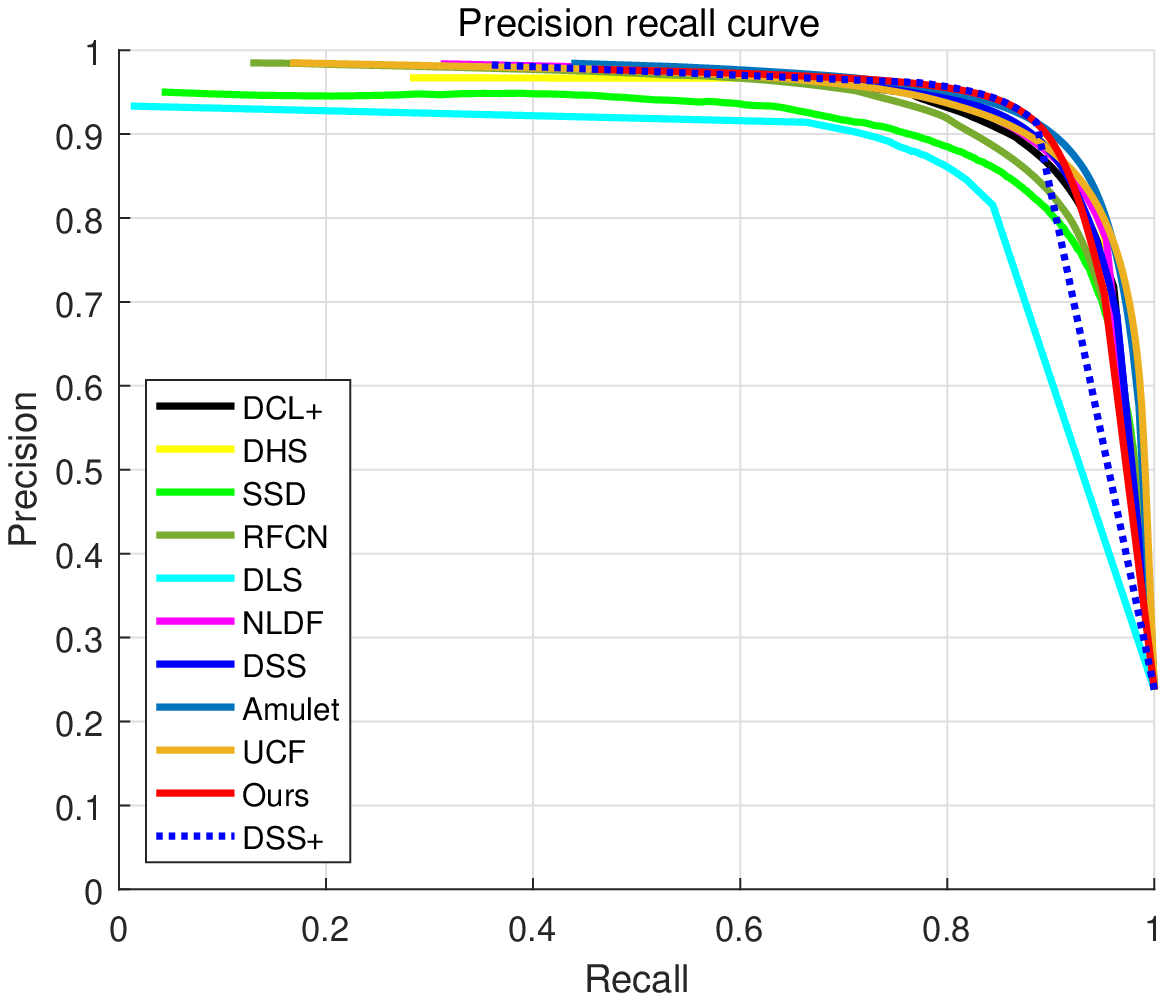}
		\caption{\scriptsize ECSSD}\label{fig_pr_ecssd}
	\end{subfigure}
	\begin{subfigure}[t]{4cm}
		\centering
		\includegraphics[width=4cm]{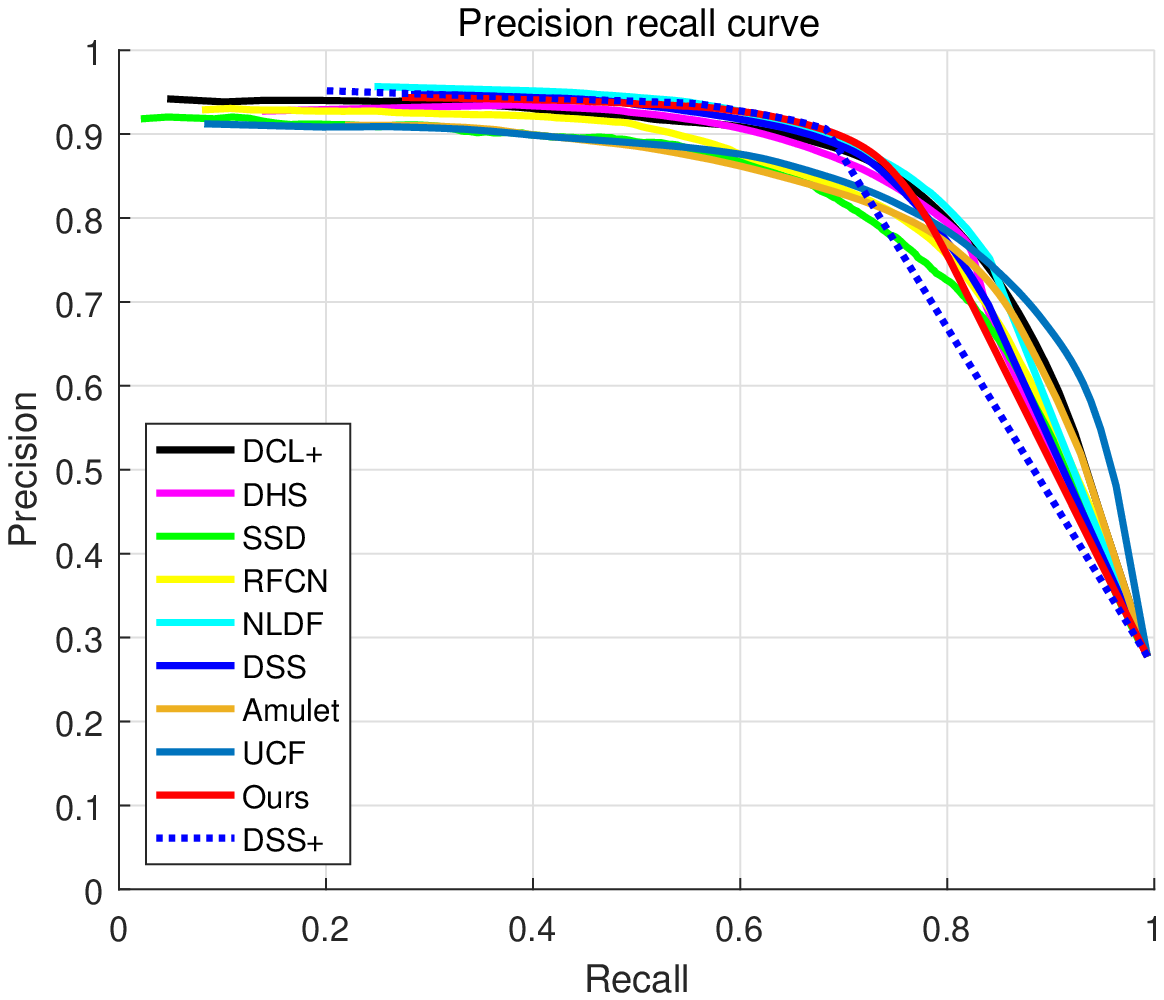}
		\caption{\scriptsize SOD}\label{fig_pr_sod}
	\end{subfigure}
        \begin{subfigure}[t]{4cm}
		\centering
		\includegraphics[width=4cm]{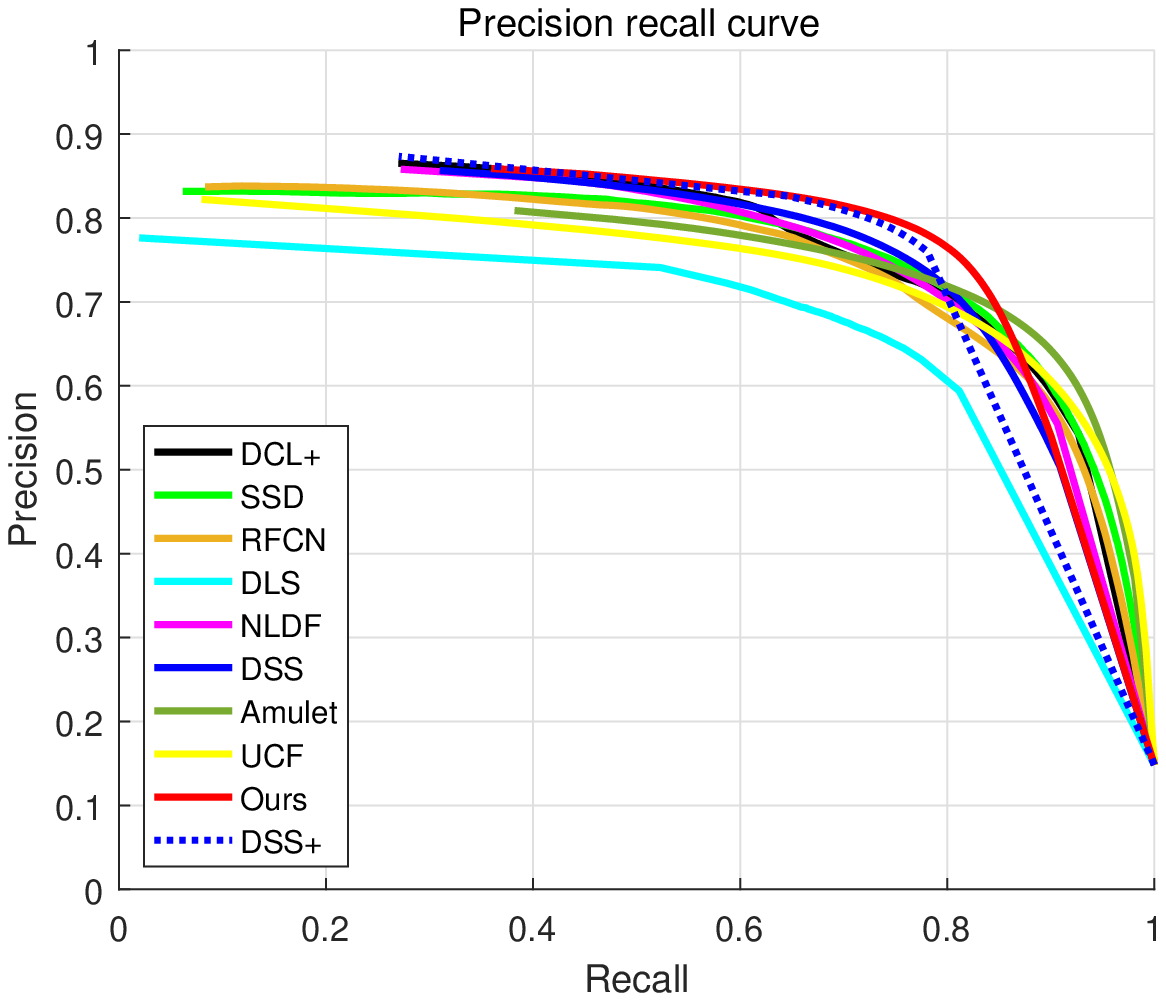}
		\caption{\scriptsize DUT-OMRON}\label{fig_pr_omron}
	\end{subfigure}
	\caption{Comparison of precision-recall curves on different datasets.}\label{fig_pr}
\end{figure}

\textbf{Quantitative Evaluation.} The results of quantitative comparison with state-of-the-art methods are reported in Table~\ref{table_results} and Fig.~\ref{fig_pr}. We can clearly observe that our approach significantly outperforms the competing methods both in terms of F-measure and MAE scores, especially on the challenging datasets (\emph{e.g.}, DUT-OMRON). For PR curves, we also achieved comparable performance with state-of-the-arts except at high level of recall (recall$>$0.9). In comparison to the top method, DSS$^{+}$, which uses a CRF-based post-processing step to refine the resolution, nevertheless, our approach still attains nearly identical (or better) performance across the board. It also needs to point out that the existing methods used different training datasets and data augmentation strategies, which caused an unfair comparison. Nevertheless, we still perform much better that clearly shows the superiority of the proposed approach. And we also believe that further performance gain can be obtained by using larger training dataset with more augmented training images, which is beyond the scope of this paper.

\begin{figure}
\captionsetup[subfigure]{labelformat=empty}	
	\centering
       \begin{subfigure}[t]{1.25cm}
		\centering
		\includegraphics[width=1.25cm]{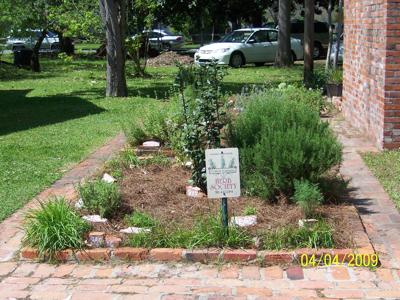}
	\end{subfigure}
	\begin{subfigure}[t]{1.25cm}
		\centering
		\includegraphics[width=1.25cm]{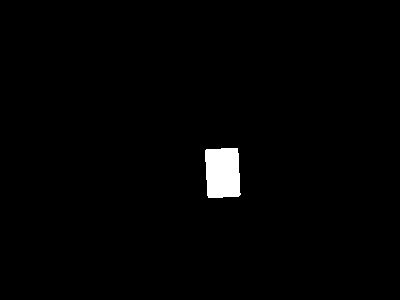}
	\end{subfigure}
	\begin{subfigure}[t]{1.25cm}
		\centering
		\includegraphics[width=1.25cm]{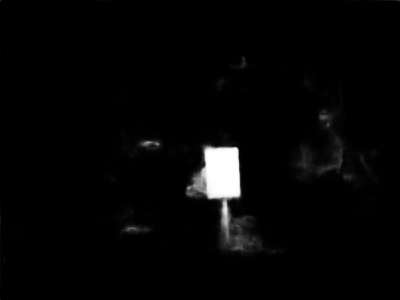}
	\end{subfigure}
	\begin{subfigure}[t]{1.25cm}
		\centering
		\includegraphics[width=1.25cm]{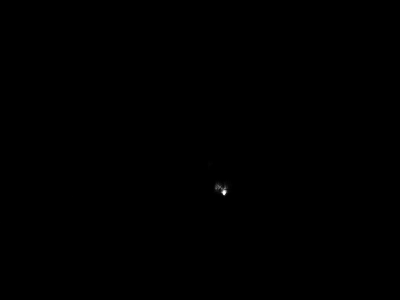}
	\end{subfigure}
	\begin{subfigure}[t]{1.25cm}
		\centering
		\includegraphics[width=1.25cm]{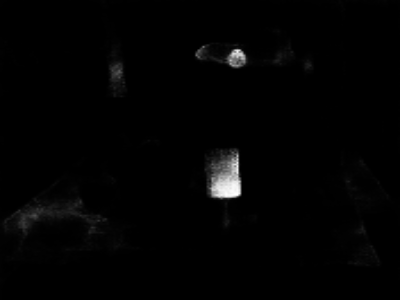}
	\end{subfigure}
	\begin{subfigure}[t]{1.25cm}
		\centering
		\includegraphics[width=1.25cm]{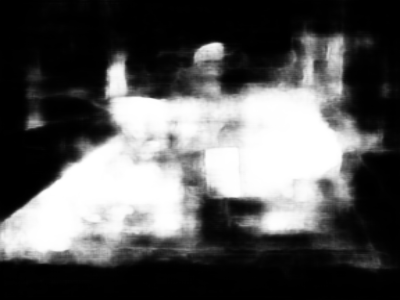}
	\end{subfigure}
	\begin{subfigure}[t]{1.25cm}
		\centering
		\includegraphics[width=1.25cm]{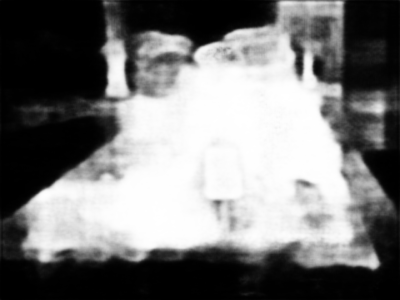}
	\end{subfigure}
	\begin{subfigure}[t]{1.25cm}
		\centering
		\includegraphics[width=1.25cm]{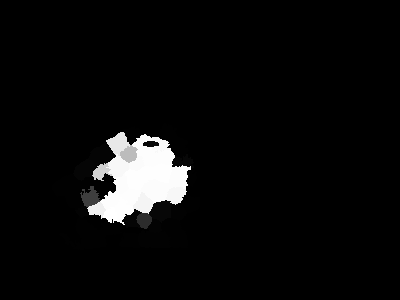}
	\end{subfigure}
        \begin{subfigure}[t]{1.25cm}
		\centering
		\includegraphics[width=1.25cm]{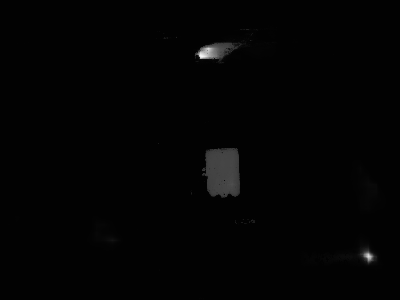}
	\end{subfigure}

      \begin{subfigure}[t]{1.25cm}
		\centering
		\includegraphics[width=1.25cm]{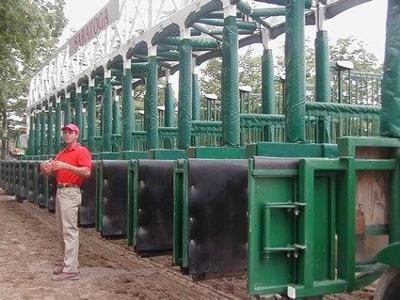}
	\end{subfigure}
	\begin{subfigure}[t]{1.25cm}
		\centering
		\includegraphics[width=1.25cm]{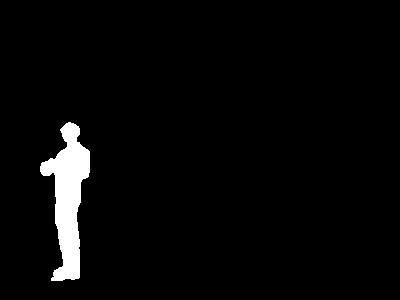}
	\end{subfigure}
	\begin{subfigure}[t]{1.25cm}
		\centering
		\includegraphics[width=1.25cm]{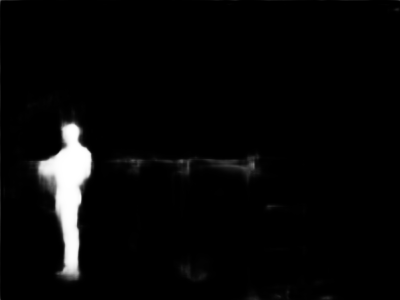}
	\end{subfigure}
	\begin{subfigure}[t]{1.25cm}
		\centering
		\includegraphics[width=1.25cm]{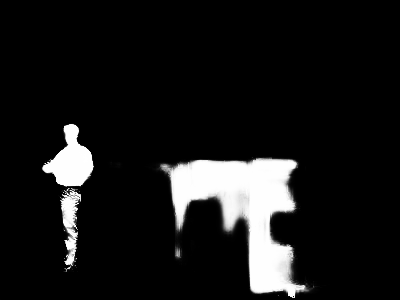}
	\end{subfigure}
	\begin{subfigure}[t]{1.25cm}
		\centering
		\includegraphics[width=1.25cm]{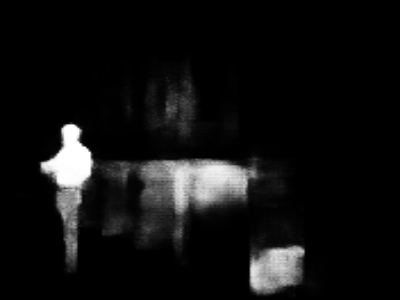}
	\end{subfigure}
	\begin{subfigure}[t]{1.25cm}
		\centering
		\includegraphics[width=1.25cm]{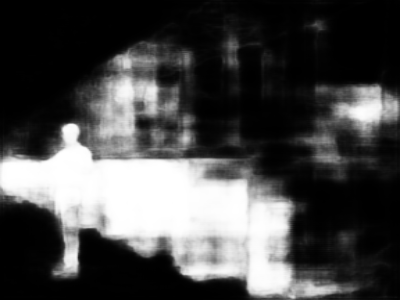}
	\end{subfigure}
	\begin{subfigure}[t]{1.25cm}
		\centering
		\includegraphics[width=1.25cm]{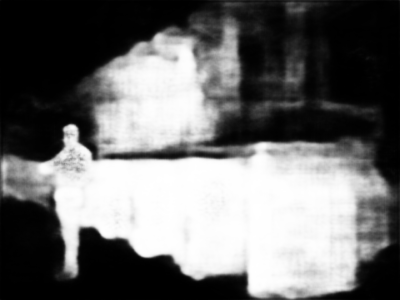}
	\end{subfigure}
	\begin{subfigure}[t]{1.25cm}
		\centering
		\includegraphics[width=1.25cm]{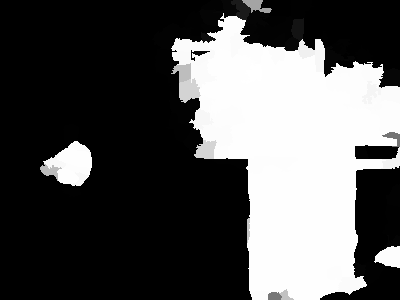}
	\end{subfigure}
        \begin{subfigure}[t]{1.25cm}
		\centering
		\includegraphics[width=1.25cm]{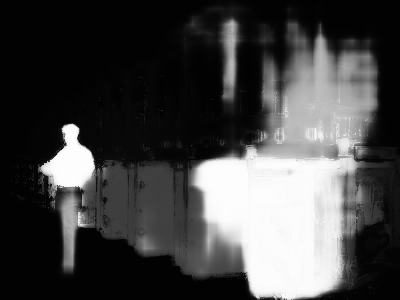}
	\end{subfigure}

       \begin{subfigure}[t]{1.25cm}
		\centering
		\includegraphics[width=1.25cm]{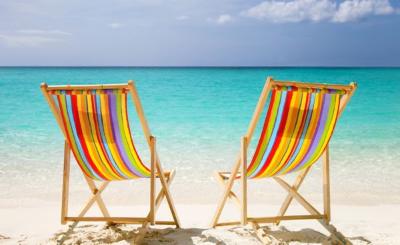}
	\end{subfigure}
	\begin{subfigure}[t]{1.25cm}
		\centering
		\includegraphics[width=1.25cm]{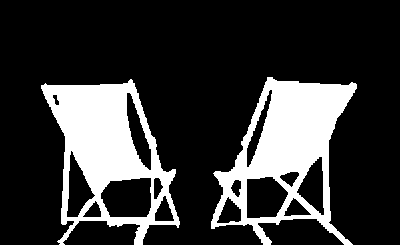}
	\end{subfigure}
	\begin{subfigure}[t]{1.25cm}
		\centering
		\includegraphics[width=1.25cm]{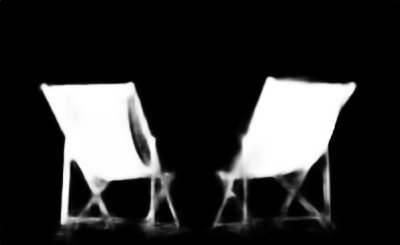}
	\end{subfigure}
	\begin{subfigure}[t]{1.25cm}
		\centering
		\includegraphics[width=1.25cm]{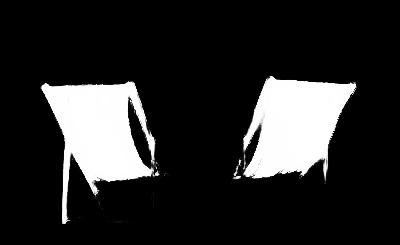}
	\end{subfigure}
	\begin{subfigure}[t]{1.25cm}
		\centering
		\includegraphics[width=1.25cm]{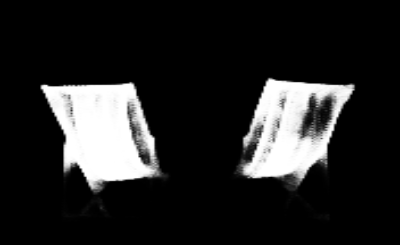}
	\end{subfigure}
	\begin{subfigure}[t]{1.25cm}
		\centering
		\includegraphics[width=1.25cm]{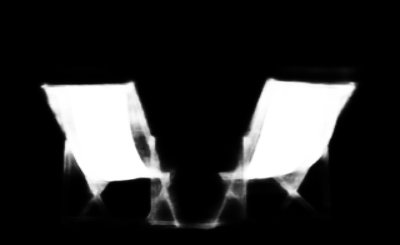}
	\end{subfigure}
	\begin{subfigure}[t]{1.25cm}
		\centering
		\includegraphics[width=1.25cm]{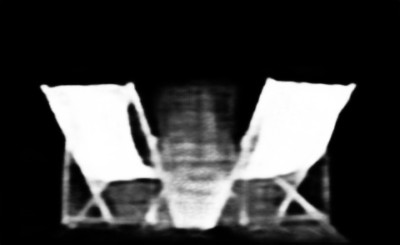}
	\end{subfigure}
	\begin{subfigure}[t]{1.25cm}
		\centering
		\includegraphics[width=1.25cm]{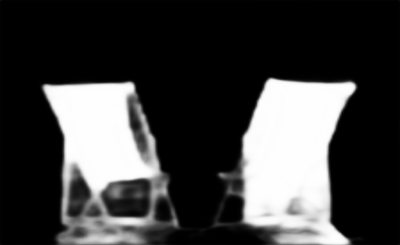}
	\end{subfigure}
        \begin{subfigure}[t]{1.25cm}
		\centering
		\includegraphics[width=1.25cm]{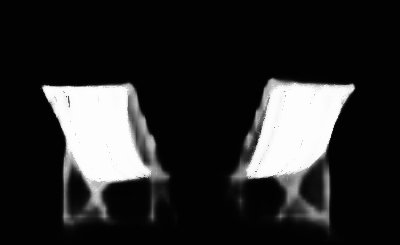}
	\end{subfigure}

       \begin{subfigure}[t]{1.25cm}
		\centering
		\includegraphics[width=1.25cm]{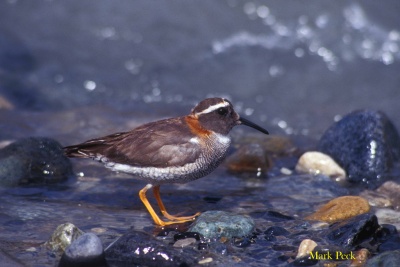}
	\end{subfigure}
	\begin{subfigure}[t]{1.25cm}
		\centering
		\includegraphics[width=1.25cm]{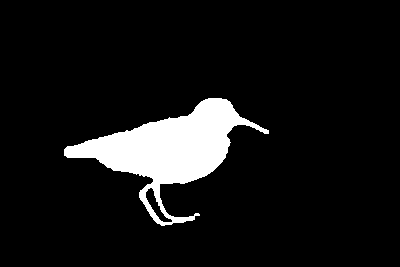}
	\end{subfigure}
	\begin{subfigure}[t]{1.25cm}
		\centering
		\includegraphics[width=1.25cm]{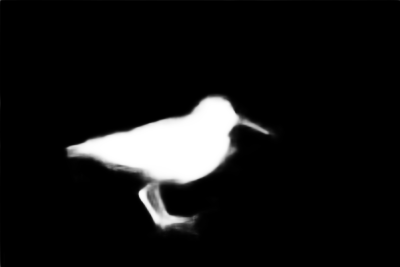}
	\end{subfigure}
	\begin{subfigure}[t]{1.25cm}
		\centering
		\includegraphics[width=1.25cm]{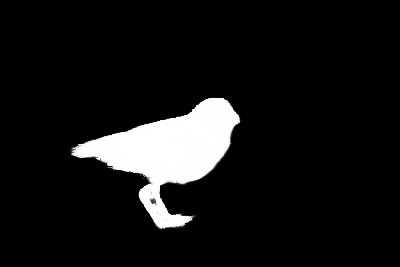}
	\end{subfigure}
	\begin{subfigure}[t]{1.25cm}
		\centering
		\includegraphics[width=1.25cm]{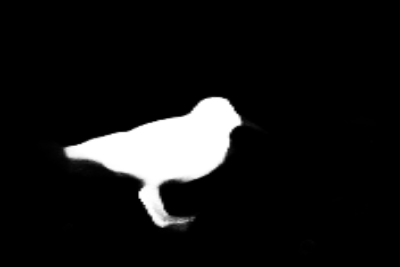}
	\end{subfigure}
	\begin{subfigure}[t]{1.25cm}
		\centering
		\includegraphics[width=1.25cm]{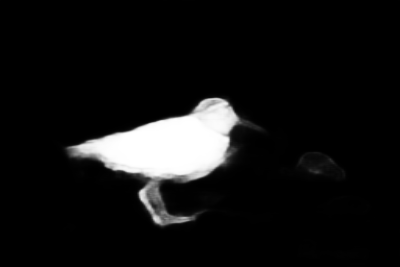}
	\end{subfigure}
	\begin{subfigure}[t]{1.25cm}
		\centering
		\includegraphics[width=1.25cm]{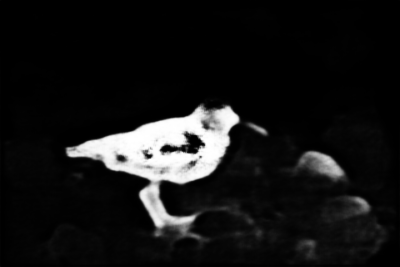}
	\end{subfigure}
	\begin{subfigure}[t]{1.25cm}
		\centering
		\includegraphics[width=1.25cm]{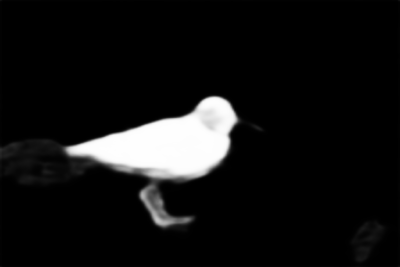}
	\end{subfigure}
        \begin{subfigure}[t]{1.25cm}
		\centering
		\includegraphics[width=1.25cm]{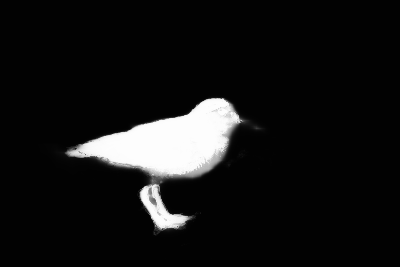}
	\end{subfigure}

       \begin{subfigure}[t]{1.25cm}
		\centering
		\includegraphics[width=1.25cm]{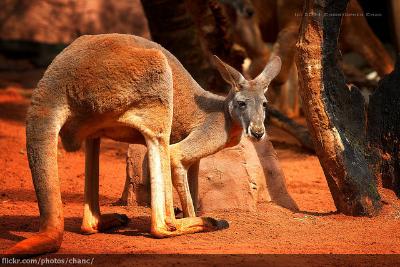}
	\end{subfigure}
	\begin{subfigure}[t]{1.25cm}
		\centering
		\includegraphics[width=1.25cm]{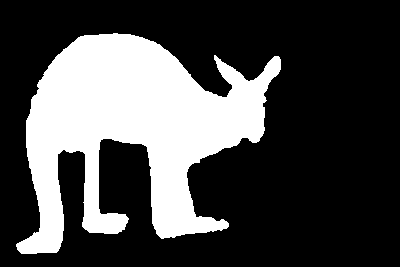}
	\end{subfigure}
	\begin{subfigure}[t]{1.25cm}
		\centering
		\includegraphics[width=1.25cm]{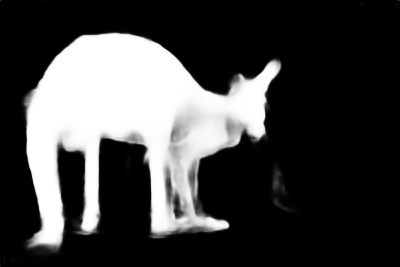}
	\end{subfigure}
	\begin{subfigure}[t]{1.25cm}
		\centering
		\includegraphics[width=1.25cm]{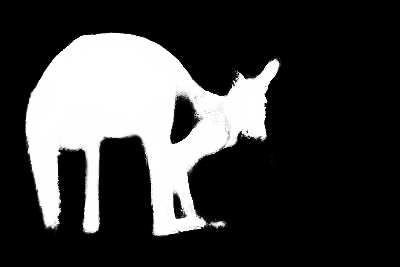}
	\end{subfigure}
	\begin{subfigure}[t]{1.25cm}
		\centering
		\includegraphics[width=1.25cm]{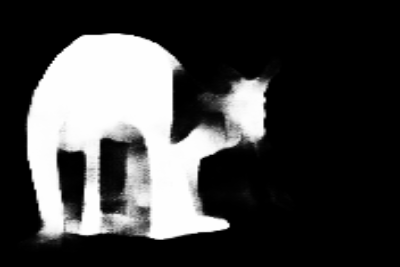}
	\end{subfigure}
	\begin{subfigure}[t]{1.25cm}
		\centering
		\includegraphics[width=1.25cm]{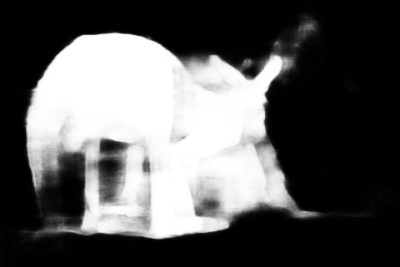}
	\end{subfigure}
	\begin{subfigure}[t]{1.25cm}
		\centering
		\includegraphics[width=1.25cm]{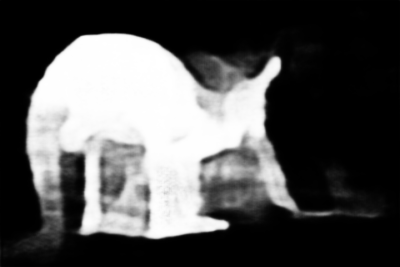}
	\end{subfigure}
	\begin{subfigure}[t]{1.25cm}
		\centering
		\includegraphics[width=1.25cm]{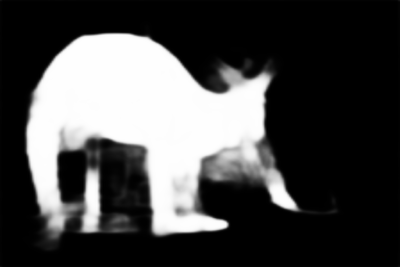}
	\end{subfigure}
        \begin{subfigure}[t]{1.25cm}
		\centering
		\includegraphics[width=1.25cm]{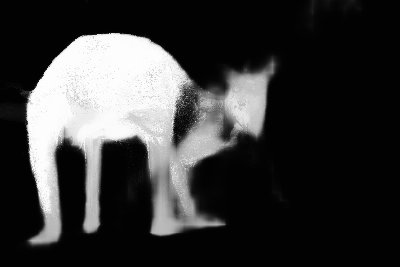}
	\end{subfigure}

      \begin{subfigure}[t]{1.25cm}
		\centering
		\includegraphics[width=1.25cm]{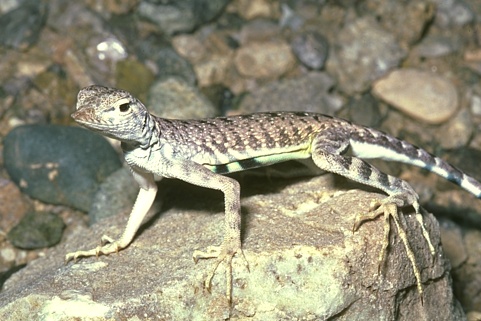}
	\end{subfigure}
	\begin{subfigure}[t]{1.25cm}
		\centering
		\includegraphics[width=1.25cm]{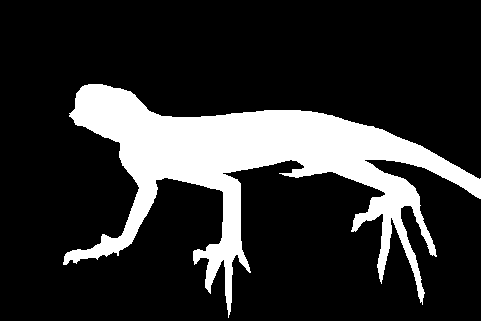}
	\end{subfigure}
	\begin{subfigure}[t]{1.25cm}
		\centering
		\includegraphics[width=1.25cm]{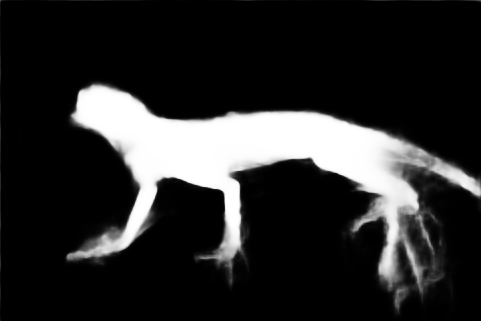}
	\end{subfigure}
	\begin{subfigure}[t]{1.25cm}
		\centering
		\includegraphics[width=1.25cm]{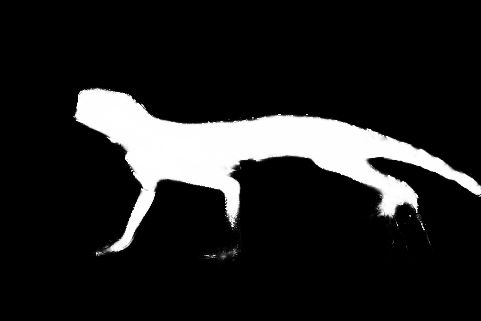}
	\end{subfigure}
	\begin{subfigure}[t]{1.25cm}
		\centering
		\includegraphics[width=1.25cm]{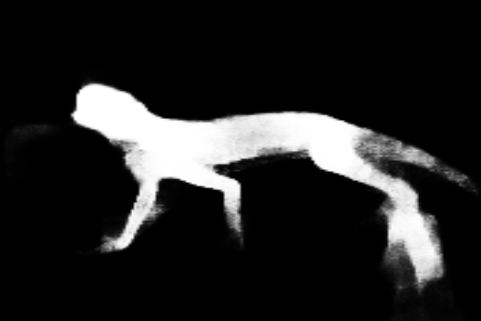}
	\end{subfigure}
	\begin{subfigure}[t]{1.25cm}
		\centering
		\includegraphics[width=1.25cm]{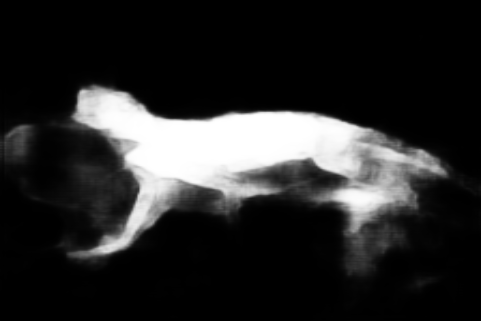}
	\end{subfigure}
	\begin{subfigure}[t]{1.25cm}
		\centering
		\includegraphics[width=1.25cm]{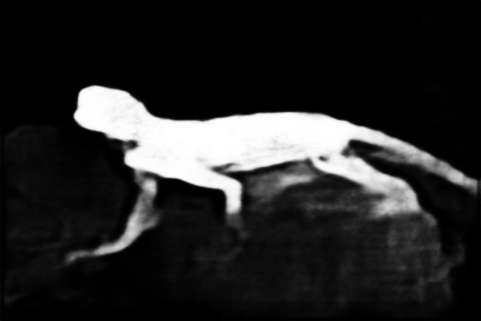}
	\end{subfigure}
	\begin{subfigure}[t]{1.25cm}
		\centering
		\includegraphics[width=1.25cm]{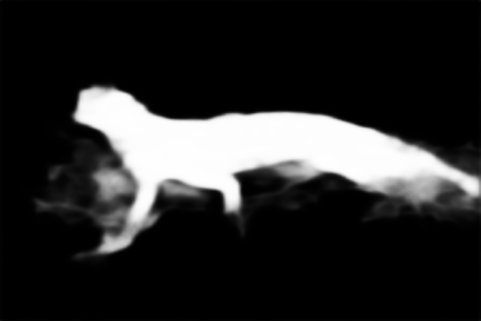}
	\end{subfigure}
        \begin{subfigure}[t]{1.25cm}
		\centering
		\includegraphics[width=1.25cm]{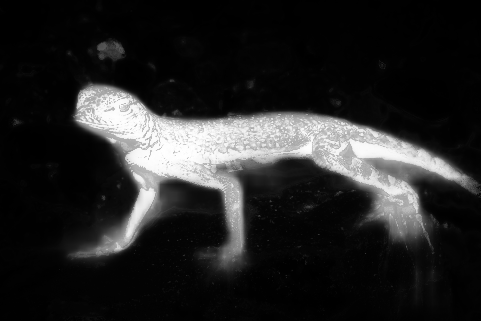}
	\end{subfigure}

       \begin{subfigure}[t]{1.25cm}
		\centering
		\includegraphics[width=1.25cm]{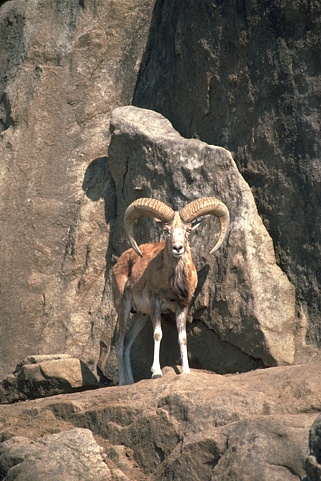}
	\end{subfigure}
	\begin{subfigure}[t]{1.25cm}
		\centering
		\includegraphics[width=1.25cm]{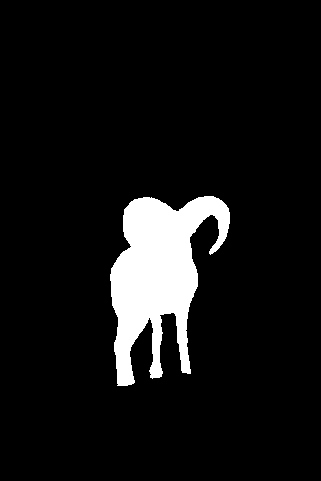}
	\end{subfigure}
	\begin{subfigure}[t]{1.25cm}
		\centering
		\includegraphics[width=1.25cm]{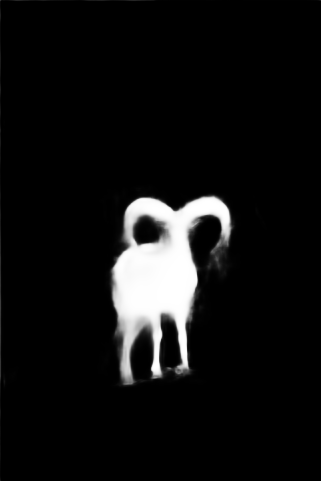}
	\end{subfigure}
	\begin{subfigure}[t]{1.25cm}
		\centering
		\includegraphics[width=1.25cm]{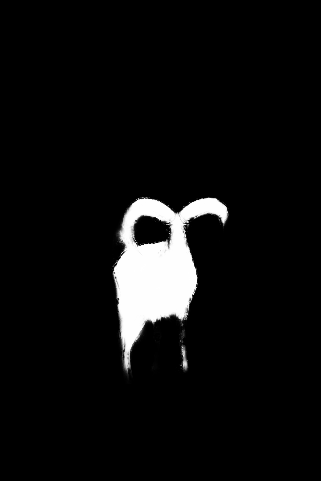}
	\end{subfigure}
	\begin{subfigure}[t]{1.25cm}
		\centering
		\includegraphics[width=1.25cm]{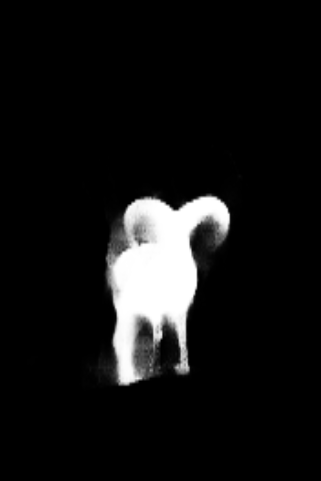}
	\end{subfigure}
	\begin{subfigure}[t]{1.25cm}
		\centering
		\includegraphics[width=1.25cm]{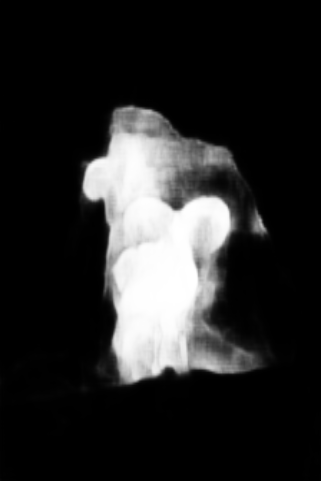}
	\end{subfigure}
	\begin{subfigure}[t]{1.25cm}
		\centering
		\includegraphics[width=1.25cm]{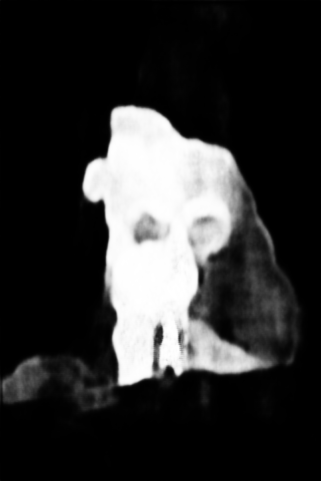}
	\end{subfigure}
	\begin{subfigure}[t]{1.25cm}
		\centering
		\includegraphics[width=1.25cm]{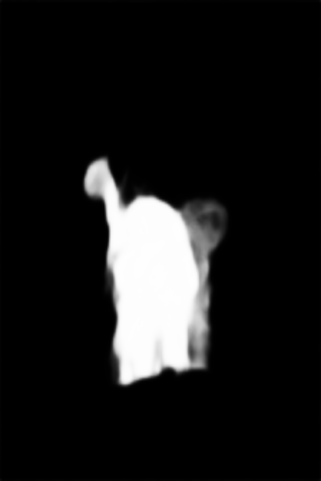}
	\end{subfigure}
        \begin{subfigure}[t]{1.25cm}
		\centering
		\includegraphics[width=1.25cm]{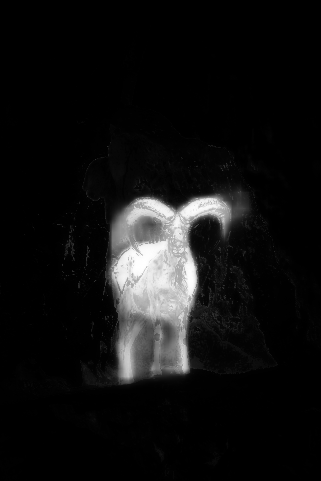}
	\end{subfigure}

       \begin{subfigure}[t]{1.25cm}
		\centering
		\includegraphics[width=1.25cm]{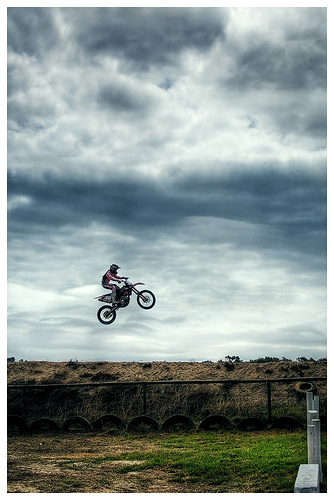}
	\end{subfigure}
	\begin{subfigure}[t]{1.25cm}
		\centering
		\includegraphics[width=1.25cm]{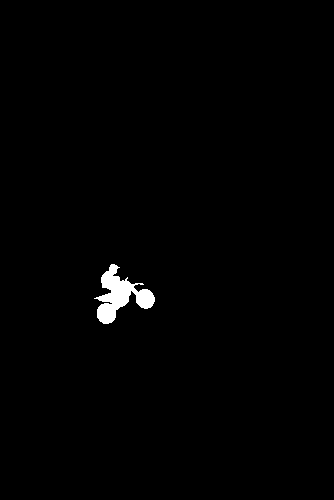}
	\end{subfigure}
	\begin{subfigure}[t]{1.25cm}
		\centering
		\includegraphics[width=1.25cm]{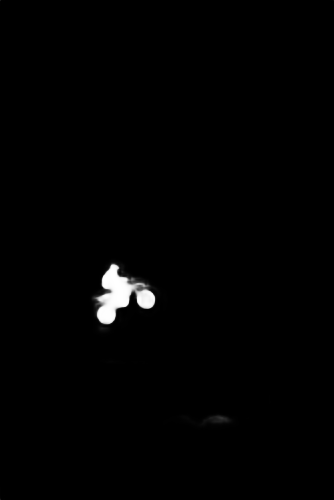}
	\end{subfigure}
	\begin{subfigure}[t]{1.25cm}
		\centering
		\includegraphics[width=1.25cm]{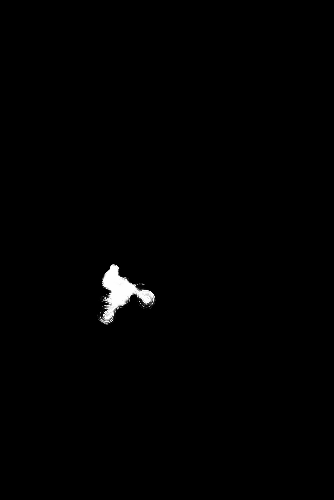}
	\end{subfigure}
	\begin{subfigure}[t]{1.25cm}
		\centering
		\includegraphics[width=1.25cm]{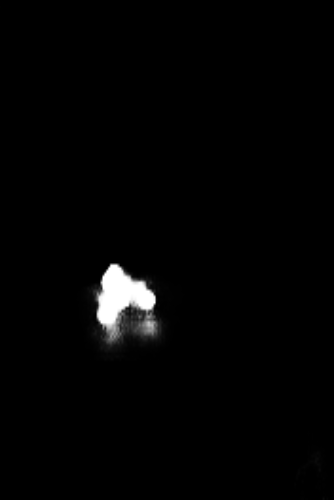}
	\end{subfigure}
	\begin{subfigure}[t]{1.25cm}
		\centering
		\includegraphics[width=1.25cm]{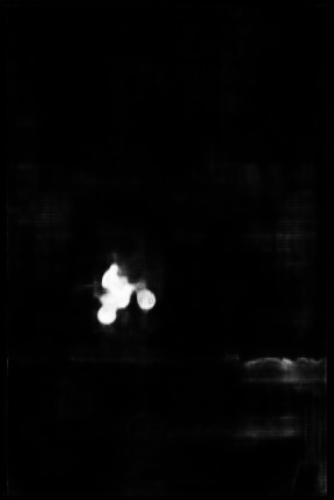}
	\end{subfigure}
	\begin{subfigure}[t]{1.25cm}
		\centering
		\includegraphics[width=1.25cm]{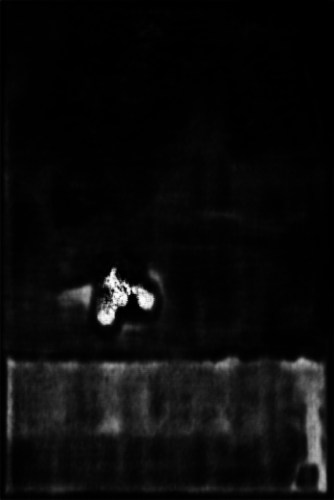}
	\end{subfigure}
	\begin{subfigure}[t]{1.25cm}
		\centering
		\includegraphics[width=1.25cm]{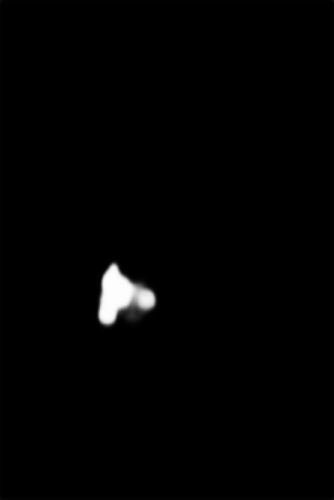}
	\end{subfigure}
        \begin{subfigure}[t]{1.25cm}
		\centering
		\includegraphics[width=1.25cm]{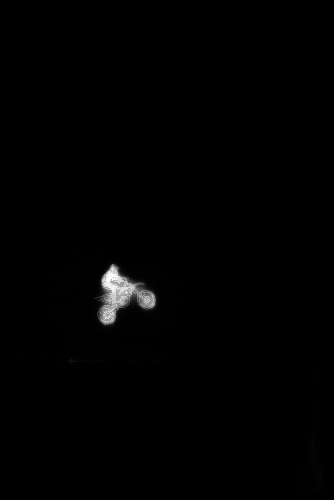}
	\end{subfigure}

       \begin{subfigure}[t]{1.25cm}
		\centering
		\includegraphics[width=1.25cm]{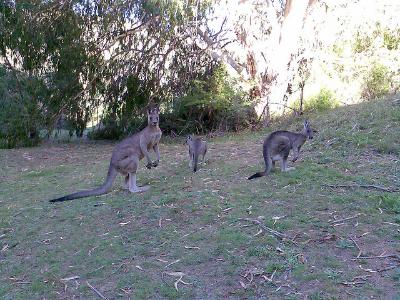}
	\end{subfigure}
	\begin{subfigure}[t]{1.25cm}
		\centering
		\includegraphics[width=1.25cm]{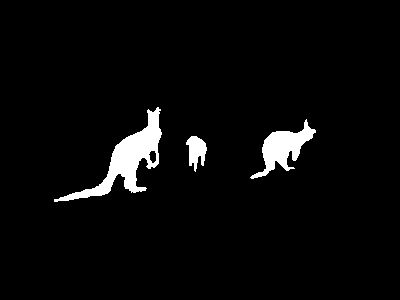}
	\end{subfigure}
	\begin{subfigure}[t]{1.25cm}
		\centering
		\includegraphics[width=1.25cm]{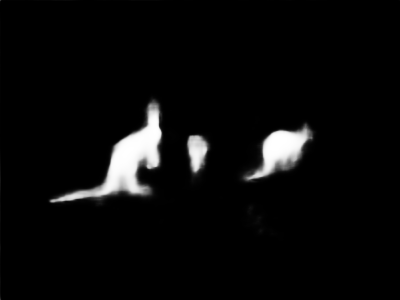}
	\end{subfigure}
	\begin{subfigure}[t]{1.25cm}
		\centering
		\includegraphics[width=1.25cm]{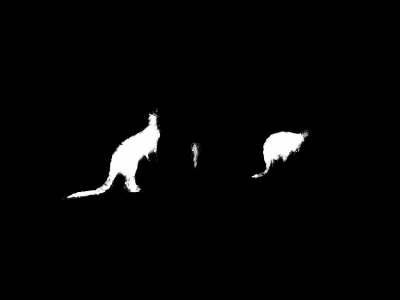}
	\end{subfigure}
	\begin{subfigure}[t]{1.25cm}
		\centering
		\includegraphics[width=1.25cm]{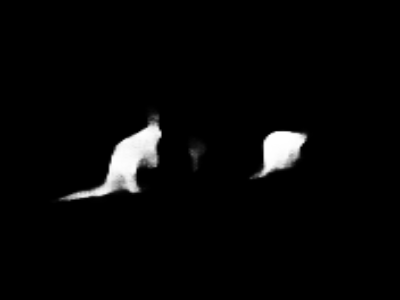}
	\end{subfigure}
	\begin{subfigure}[t]{1.25cm}
		\centering
		\includegraphics[width=1.25cm]{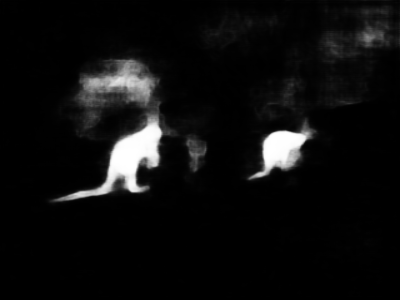}
	\end{subfigure}
	\begin{subfigure}[t]{1.25cm}
		\centering
		\includegraphics[width=1.25cm]{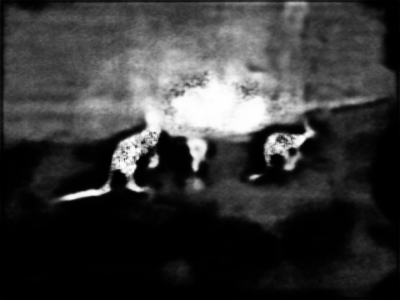}
	\end{subfigure}
	\begin{subfigure}[t]{1.25cm}
		\centering
		\includegraphics[width=1.25cm]{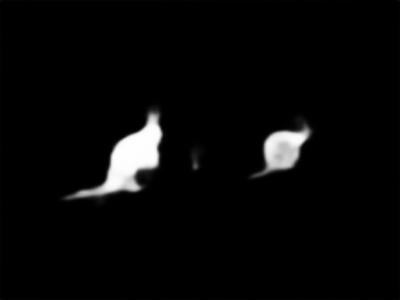}
	\end{subfigure}
        \begin{subfigure}[t]{1.25cm}
		\centering
		\includegraphics[width=1.25cm]{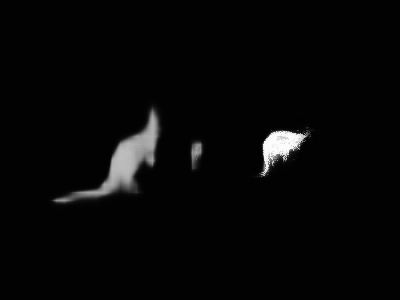}
	\end{subfigure}

       \begin{subfigure}[t]{1.25cm}
		\centering
		\includegraphics[width=1.25cm]{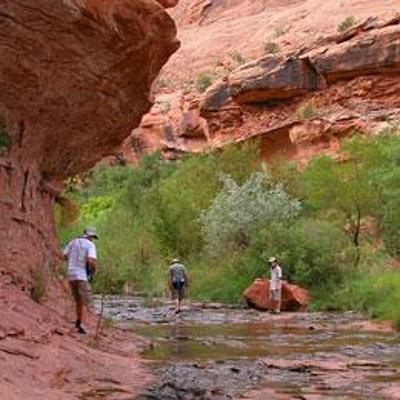}
	\end{subfigure}
	\begin{subfigure}[t]{1.25cm}
		\centering
		\includegraphics[width=1.25cm]{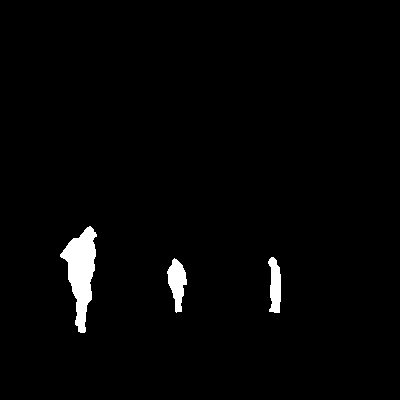}
	\end{subfigure}
	\begin{subfigure}[t]{1.25cm}
		\centering
		\includegraphics[width=1.25cm]{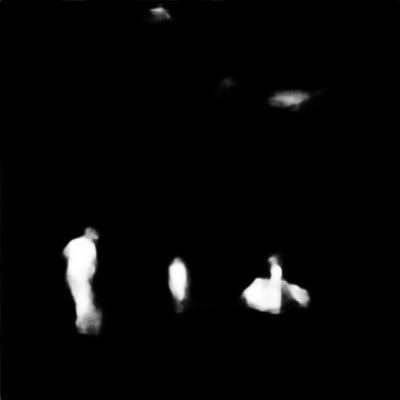}
	\end{subfigure}
	\begin{subfigure}[t]{1.25cm}
		\centering
		\includegraphics[width=1.25cm]{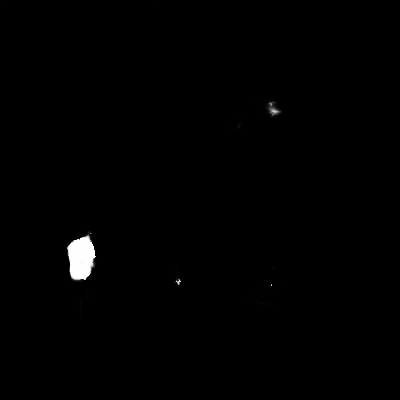}
	\end{subfigure}
	\begin{subfigure}[t]{1.25cm}
		\centering
		\includegraphics[width=1.25cm]{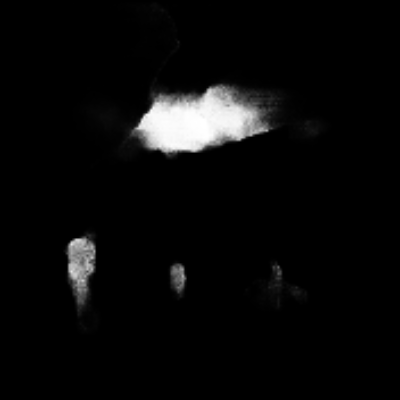}
	\end{subfigure}
	\begin{subfigure}[t]{1.25cm}
		\centering
		\includegraphics[width=1.25cm]{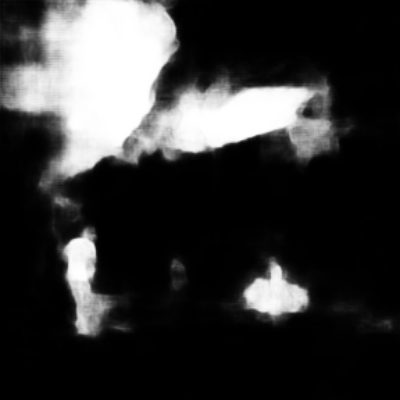}
	\end{subfigure}
	\begin{subfigure}[t]{1.25cm}
		\centering
		\includegraphics[width=1.25cm]{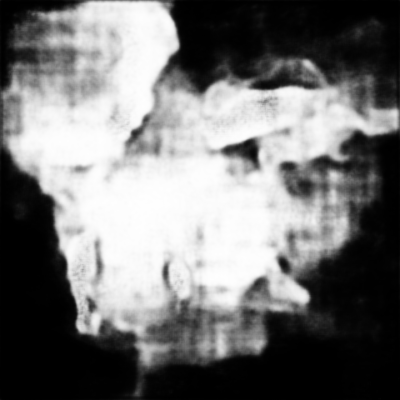}
	\end{subfigure}
	\begin{subfigure}[t]{1.25cm}
		\centering
		\includegraphics[width=1.25cm]{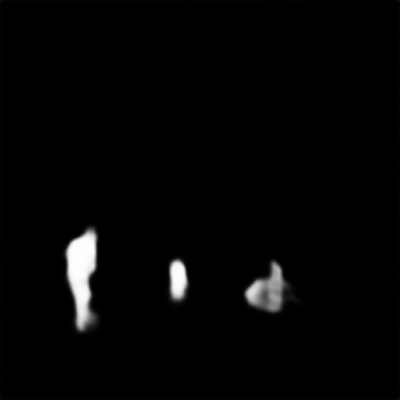}
	\end{subfigure}
        \begin{subfigure}[t]{1.25cm}
		\centering
		\includegraphics[width=1.25cm]{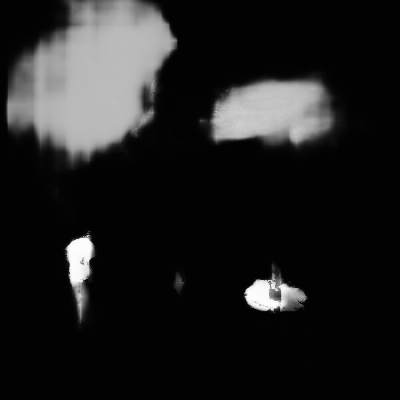}
	\end{subfigure}
                
       \begin{subfigure}[t]{1.25cm}
		\centering
		\includegraphics[width=1.25cm]{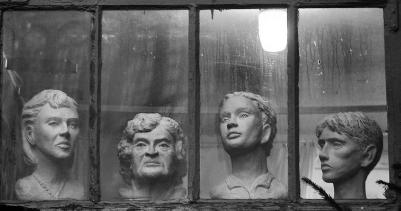}
		\caption{\scriptsize Img}	
	\end{subfigure}
	\begin{subfigure}[t]{1.25cm}
		\centering
		\includegraphics[width=1.25cm]{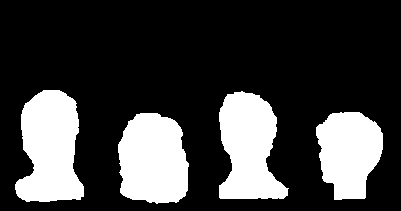}
		\caption{\scriptsize GT}	
	\end{subfigure}
	\begin{subfigure}[t]{1.25cm}
		\centering
		\includegraphics[width=1.25cm]{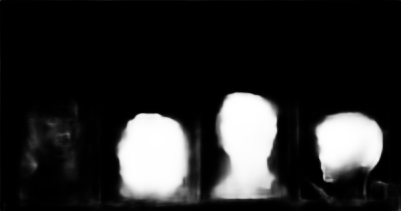}
		\caption{\scriptsize Ours}	
	\end{subfigure}
	\begin{subfigure}[t]{1.25cm}
		\centering
		\includegraphics[width=1.25cm]{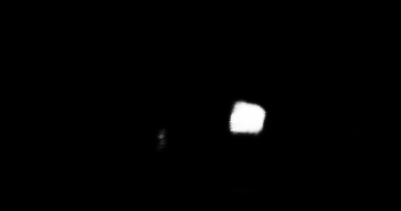}
		\caption{\scriptsize DSS$^{+}$\cite{Hou10}}	
	\end{subfigure}
	\begin{subfigure}[t]{1.25cm}
		\centering
		\includegraphics[width=1.25cm]{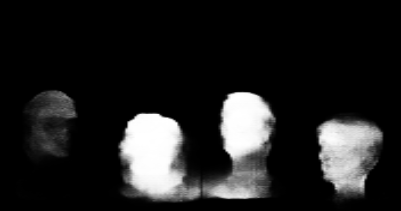}
		\caption{\scriptsize NLDF\cite{Luo22}}	
	\end{subfigure}
	\begin{subfigure}[t]{1.25cm}
		\centering
		\includegraphics[width=1.25cm]{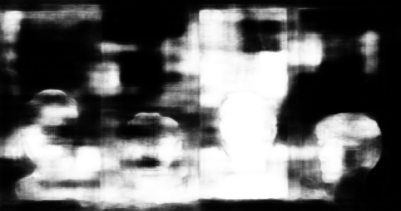}
		\caption{\scriptsize Amulet\cite{Zhang14}}	
	\end{subfigure}
	\begin{subfigure}[t]{1.25cm}
		\centering
		\includegraphics[width=1.25cm]{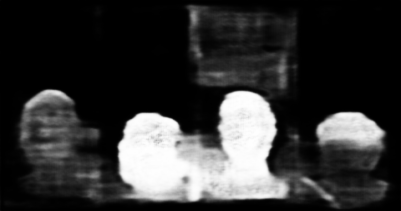}
		\caption{\scriptsize UCF\cite{Zhang23}}	
	\end{subfigure}
	\begin{subfigure}[t]{1.25cm}
		\centering
		\includegraphics[width=1.25cm]{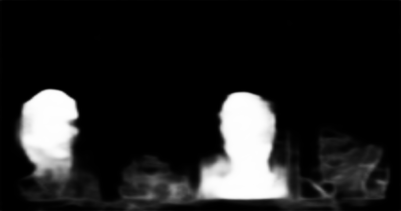}
		\caption{\scriptsize DHS\cite{Liu39}}	
	\end{subfigure}
        \begin{subfigure}[t]{1.25cm}
		\centering
		\includegraphics[width=1.25cm]{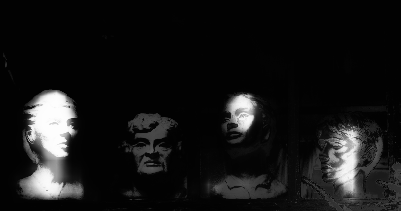}
		\caption{\scriptsize DCL$^{+}$\cite{Li6}}	
	\end{subfigure}							
	\caption{Visual comparisons with the existing methods in some challenging cases:  complex scenes, low contrast, and multiple (small) salient objects.}\label{fig_smaps}
\end{figure}

\textbf{Qualitative Evaluation.} We also show some visual results of some representative images to exhibit the superiority of the proposed approach in Fig.~\ref{fig_smaps}, including complex scenes, low contrast between salient object and background, multiple (small) salient objects with diverse characteristics (\emph{e.g.}, size, color). Taking all the cases into account, it can be observed clearly that our approach not only highlights the salient regions correctly with less false detection but also produces sharp boundaries and coherent details (\emph{e.g.}, the \emph{mouth} of the bird in the 4th row of Fig.~\ref{fig_smaps}). It is also interesting to note that the proposed method even corrected some false labeling in the ground truth, \emph{e.g.}, the left \emph{horn} in the 7th row of Fig.~\ref{fig_smaps}. Nevertheless, we still obtain unsatisfactory results in some challenging cases, taking the last row of Fig.~\ref{fig_smaps} for example, to segment all the salient objects completely is still very difficult for the existing methods.

\begin{table}[]
\centering
\caption{Quantitative comparison with state-of-the-art methods on six benchmark datasets. Each cell (from up to down) contains max F-measure (higher better), and MAE (lower better). The top two results are highlighted in {\color{red} \textbf{red}} and \textcolor[rgb]{0.13,0.55,0.13}{\textbf{green}} respectively. ``RA'' denotes the proposed reverse attention, and ``MK'' is MSRA-10K~\cite{Cheng43}, the other abbreviations are the initials of each dataset metioned in the paper. Note that the number of images listed here are including the augmented ones.}
\label{table_results}
\begin{tabular}{ccccccccc}
\hline
                                & \multicolumn{2}{c}{Training} & \multirow{2}{*}{MSRA-B} & \multirow{2}{*}{HKU-IS} &          \multirow{2}{*}{ECSSD} & \multirow{2}{*}{PASCAL-S}  & \multirow{2}{*}{SOD}  & DUT- \\ \cline{2-3}  & Dataset & \#Images & & & & & & OMRON \\\hline
\multirow{2}{*}{DRFI~\cite{Jiang37}} & \multirow{2}{*}{MB} & \multirow{2}{*}{2.5k} & 0.851  & 0.775  & 0.784 & 0.690  & 0.699 & 0.664  \\
                                & & & 0.123  & 0.146  & 0.172 & 0.210 & 0.223 & 0.150     \\ \hline
\multirow{2}{*}{DCL$^{+}$~\cite{Li6}} & \multirow{2}{*}{MB} & \multirow{2}{*}{2.5k} & 0.918  & 0.907  & 0.898 & 0.810  & 0.831 & 0.757  \\
                                & & & 0.047  & 0.048  & 0.071 & 0.115 & 0.131 & 0.080     \\ \hline
\multirow{2}{*}{DHS\cite{Liu39}}  & \multirow{2}{*}{MK+D}  & \multirow{2}{*}{9.5k$\times$12}  & -   & 0.892  & 0.905 & 0.824  & 0.823 & -    \\
                                & & & -    & 0.052  & 0.061 & \color{red} \textbf{0.094}  & 0.127 & -         \\ \hline
\multirow{2}{*}{SSD\cite{Kim40}} & \multirow{2}{*}{MB} & \multirow{2}{*}{2.5k} & 0.902  & -   & 0.865 & 0.774  & 0.793 & 0.754  \\
                                & & & 0.160  & -    & 0.193 & 0.220  & 0.222 & 0.193     \\ \hline
\multirow{2}{*}{RFCN\cite{Wang21}} & \multirow{2}{*}{MK} & \multirow{2}{*}{10k} & -   & 0.894  & 0.889 & 0.829  & 0.799 & 0.744  \\
                                & & & -      & 0.088  & 0.109 & 0.133  & 0.169 & 0.111     \\ \hline
\multirow{2}{*}{DLS\cite{Hu12}} & \multirow{2}{*}{MK} & \multirow{2}{*}{10k} & -      & 0.835  & 0.852 & 0.753  & -     & 0.687  \\
                                & & & -      & 0.070  & 0.088 & 0.132   & -     & 0.090     \\ \hline
\multirow{2}{*}{NLDF\cite{Luo22}} & \multirow{2}{*}{MB}  & \multirow{2}{*}{2.5k$\times$2} & 0.911  & 0.902  & 0.903 & 0.826 & 0.837 & 0.753 \\
                                & & & 0.048  & 0.048  & 0.065 & 0.099 & 0.123 & 0.080     \\ \hline
\multirow{2}{*}{Amulet\cite{Zhang14}} & \multirow{2}{*}{MK} & \multirow{2}{*}{10k$\times$8} & -  & 0.899  & 0.914 & 0.832 & 0.795 & 0.743 \\
                                & & & -  & 0.050  & 0.061 & 0.100  & 0.144 & 0.098     \\ \hline
\multirow{2}{*}{UCF\cite{Zhang23}} & \multirow{2}{*}{MK} & \multirow{2}{*}{10k$\times$8} & -  & 0.888  & 0.902 & 0.818 & 0.805 & 0.730 \\
                                & & & -   & 0.061  & 0.071 & 0.116 & 0.148 & 0.120     \\ \hline                                
\multirow{2}{*}{DSS\cite{Hou10}} & \multirow{2}{*}{MB} & \multirow{2}{*}{2.5k$\times$2} & 0.920  & 0.900  & 0.908 & 0.826 & 0.834 & 0.764  \\
                                & & & 0.043  & 0.050  & 0.063 & 0.102 & 0.126 & 0.072     \\\hline
\multirow{2}{*}{DSS$^{+}$\cite{Hou10}} & \multirow{2}{*}{MB} & \multirow{2}{*}{2.5k$\times$2} & \textcolor[rgb]{0.13,0.55,0.13}{\textbf{0.929}}  & \color{red} \textbf{0.916}  & \color{red} \textbf{0.919} & \color{red} \textbf{0.835} & \textcolor[rgb]{0.13,0.55,0.13}{\textbf{0.843}} & \textcolor[rgb]{0.13,0.55,0.13}{\textbf{0.781}}  \\
                                & & & \color{red} \textbf{0.034}  & \color{red} \textbf{0.040}  & \color{red} \textbf{0.055} & \textcolor[rgb]{0.13,0.55,0.13}{\textbf{0.095}} & \color{red} \textbf{0.122} & \textcolor[rgb]{0.13,0.55,0.13}{\textbf{0.063}}     \\\hline
Ours & \multirow{2}{*}{MB} & \multirow{2}{*}{2.5k$\times$2} & 0.919  & 0.898  & 0.905 & 0.818 & 0.839 & 0.762  \\
w/o RA & & & 0.042  & 0.049  & 0.063 & 0.106 & 0.126 & 0.071 \\\hline
\multirow{2}{*}{Ours} & \multirow{2}{*}{MB} & \multirow{2}{*}{2.5k$\times$2} & \color{red} \textbf{0.931}  & \textcolor[rgb]{0.13,0.55,0.13}{\textbf{0.913}}  & \textcolor[rgb]{0.13,0.55,0.13}{\textbf{0.918}} & \textcolor[rgb]{0.13,0.55,0.13}{\textbf{0.834}} & \color{red} \textbf{0.844} & \color{red} \textbf{0.786}  \\
                                & & & \textcolor[rgb]{0.13,0.55,0.13}{\textbf{0.036}}  & \textcolor[rgb]{0.13,0.55,0.13}{\textbf{0.045}}  & \textcolor[rgb]{0.13,0.55,0.13}{\textbf{0.059}} & 0.104 & \textcolor[rgb]{0.13,0.55,0.13}{\textbf{0.124}} & \color{red} \textbf{0.062} \\
\hline                                
\end{tabular}
\end{table}

\textbf{Execution Time.} Finally, we investigate the efficiency of our method, and conduct all the experiments on a single NVIDIA TITAN Xp GPU for fair comparison. It only takes less than 2 hours to train our model, for comparison, DSS needs about 6 hours. We also compared the average execution time with other five leading CNN-based methods on ECSSD. As can be seen from Table~\ref{table_time}, our approach is much faster than all the competing methods. Therefore, considering both in vusial quality and efficiency, our approach is the best choice for real-time applications up to now.

\begin{table}[]
\centering
\caption{Average execution time comparison with other methods on ECSSD.}
\label{table_time}
\begin{tabular}{p{1.3cm}<{\centering} p{1.3cm}<{\centering} p{1.3cm}<{\centering} p{1.3cm}<{\centering} p{1.3cm}<{\centering} p{1.3cm}<{\centering} p{1.3cm}<{\centering}}
\hline
         & DHS   & DSS   & NLDF  & UCF   & Amulet  & \textbf{Ours}  \\ \hline
Times(s) & 0.026 & 0.048 & 0.048 & 0.168 & 0.080  & \textbf{0.022} \\ \hline
\end{tabular}
\end{table}

\section{Conclusions}

As a low-level pre-processing step, salient object detection has great applicability in various high-level tasks yet remains not being well solved, which mainly lies on the following two aspects: low resolution output and heavy model weight. In this paper, we presented an accurate yet compact deep network for efficient salient object detection. Instead of directly learning multi-scale saliency features in different side-output stages, we employ residual learning to learn side-output residual features for saliency refinement. Based on it, the resolution of the global saliency map generated by the deepest convolutional layer was improved gradually with very limited parameters. We further propose reverse attention to guide such side-output residual learning in a top-down manner. Benefit from it, our network learned more accurate residual features, which leads to significant performance improvement. Extensive experimental results demonstrate that the proposed approach performs favorably against state-of-the-art ones both in quantitative and qualitative comparisons, which enables it a better choice for further real-world applications, and also makes it a great potential to apply in other end-to-end pixel-level prediction tasks. Nevertheless, the global saliency branch and backbone (VGG-16) network still contain large redundancy, which will be further explored by introducing handcrafted saliency prior and learning from scratch in our future work.

\section*{Acknowledgment}
This work was supported by the Natural Science Foundation of China (No. 61502412), Natural Science Foundation for Youths of Jiangsu Province (No. BK20150459), Foundation of Yangzhou University (No. 2017CXJ026).

\bibliographystyle{splncs04}
\bibliography{0356}


\end{document}